\documentclass[lettersize,journal]{IEEEtran}
\usepackage{amsmath,amsfonts}
\usepackage{algorithmic}
\usepackage{array}
\usepackage[caption=false,font=normalsize,labelfont=sf,textfont=sf]{subfig}
\usepackage{textcomp}
\usepackage{stfloats}
\usepackage{url}
\usepackage{verbatim}
\usepackage{graphicx}
\def\BibTeX{{\rm B\kern-.05em{\sc i\kern-.025em b}\kern-.08em
    T\kern-.1667em\lower.7ex\hbox{E}\kern-.125emX}}
\usepackage{balance}

\usepackage{times}
\usepackage{soul}
\usepackage{url}
\usepackage[hidelinks]{hyperref}
\usepackage[utf8]{inputenc}
\usepackage[small]{caption}
\usepackage{graphicx}
\usepackage{amsmath}
\usepackage{amsthm}
\usepackage{booktabs}
\usepackage{algorithm}
\usepackage{algorithmic}
\usepackage[switch]{lineno}
\usepackage{amsfonts,amssymb}
\usepackage{stfloats}
\usepackage{multirow}
\usepackage{graphicx}
\usepackage{cite}
\usepackage[numbers,sort&compress]{natbib}

\usepackage{colortbl}
\definecolor{mycolor}{RGB}{217,217,217}

\usepackage{pifont}       
\usepackage{bbding}       
\usepackage{fontawesome}  
 
\newcommand{\cmark}{\ding{51}}
\newcommand{\xmark}{\ding{55}}

\begin{document}
\title{Dual Frequency Branch Framework with Reconstructed Sliding Windows Attention for AI-Generated Image Detection}
\author{Jiazhen Yan, Ziqiang Li, Fan Wang, Ziwen He, Zhangjie Fu,~\IEEEmembership{Member,~IEEE,}
\thanks{This work was supported in part by the National Natural Science Foundation of China under Grant U22B2062, 62172232,62502215, the General Program of Natural Science Research in Universities of Jiangsu under Grants 25KJB520027, and Jiangsu Provincial Science and Technology Major Project (No. BG2024042). (Corresponding author: Zhangjie Fu).}
\thanks{Jiazhen Yan, Ziqiang Li, Fan Wang, Ziwen He and Zhangjie Fu are with the Engineering Research Center of Digital Forensics, Ministry of Education, Nanjing University of Information Science and Technology, Nanjing, 210044, China. (e-mail: 247918horizon@gmail.com, iceli@mail.ustc.edu.cn, wf71103@126.com, \{ziwen.he, fzj\}@nuist.edu.cn).}
}

\markboth{Journal of \LaTeX\ Class Files,~Vol.~18, No.~9, September~2020}%
{How to Use the IEEEtran \LaTeX \ Templates}

\maketitle

\begin{abstract}

    The rapid advancement of Generative Adversarial Networks (GANs) and diffusion models has enabled the creation of highly realistic synthetic images, presenting significant societal risks, such as misinformation and deception. As a result, detecting AI-generated images has emerged as a critical challenge. Existing research emphasizes extracting fine-grained features to enhance detector generalization, yet they often lack consideration for the importance and interdependencies of internal elements within local regions and are limited to a single frequency domain, hindering the capture of general forgery traces. 
    To overcome the aforementioned limitations, we first utilize a sliding window to restrict the attention mechanism to a local window, and  reconstruct the features within the window to model the relationships between neighboring internal elements within the local region. Then, we design a dual frequency domain branch framework consisting of four frequency domain subbands of DWT and the phase part of FFT to enrich the extraction of local forgery features from different perspectives. Through  feature enrichment of dual frequency domain branches and fine-grained feature extraction of reconstruction sliding window attention, our method achieves superior generalization detection capabilities on both GAN and diffusion model-based generative images.
    Evaluated on diverse datasets comprising images from 65 distinct generative models, our approach achieves a 2.13\% improvement in detection accuracy over state-of-the-art methods. Our code is available at \url{https://github.com/HorizonTEL/DFFreq-main}.

\end{abstract}


\section{Introduction}

\IEEEPARstart{W}{ith} the rapid advancement of deep learning, generative models \cite{karras2019style,dhariwal2021diffusion,li2025peer} such as GANs and diffusion models have revolutionized content creation, finding widespread applications across industries including entertainment and media. However, this progress has also introduced significant challenges \cite{wang2025pair,tao2025sagnet,tao2025oddn}. The increasing sophistication of deepfake technology has allowed the generation of highly realistic fake images and videos, making them increasingly difficult to detect and posing serious risks to societal trust, individual privacy, and security \cite{li2024proxy,li2025explore,wang2026map,wang2026spurious}. 

To prevent the abuse of generative methods, a series of AI-generated image detection mechanisms have been conceived. Early methods primarily focused on analyzing image features, such as texture, color, and lighting \cite{yu2020mining,liu2020global} to identify forgery traces. However, these methods are usually limited to the domain consistent scenario, where the training and testing sets come from the same type of generative method, resulting in poor generalized detection performance in cross-generative method scenarios.
With the continuous deepening of research, it is widely recognized that artifacts are often confined to subtle and localized areas. Inspired by this, extensive research has been conducted to extract generalizable and fine-grained forgery features \cite{tan2023learning,ojha2023towards,tan2024frequency,tan2025c2p,yan2024orthogonal,yan2024sanity,chen2024drct,yan2025ns,guillaro2025bias,pellegrini2025ai,chen2025dual}. Specifically, Tan \textit{et al.} \cite{tan2024rethinking} extracted more generalized artifact information from the generator’s perspective by analyzing the relationship between adjacent pixels. Other works \cite{chen2021local,baraldi2025contrasting,zheng2024breaking} have further amplified the local artifacts by means of segmenting patches, thus achieving better detection performance. In addition, some studies \cite{jeong2022frepgan,tan2023learning,tan2024frequency} captured forgery traces from the perspective of frequency domain to obtain more generalized detection performance. 

Although extracting local features has been widely adopted to capture generalized forgery artifacts, there are still some deficiencies as follow. 
1) The existing works all lack consideration of the importance differences and interdependencies of internal elements of local regions. Although the common attention mechanism \cite{xu2015show,vaswani2017attention,dosovitskiy2020image} can adaptively assign different weights to different local regions, they are not sensitive enough to the importance and interrelationship among the internal elements within the local regions. 2) Most existing works are limited to extracting local artifact features in a single frequency domain, resulting in the extracted features being insufficient for precisely capturing the general forgery traces caused by diverse generation methods.

\begin{figure}
    \centering
    \includegraphics[width=1\linewidth]{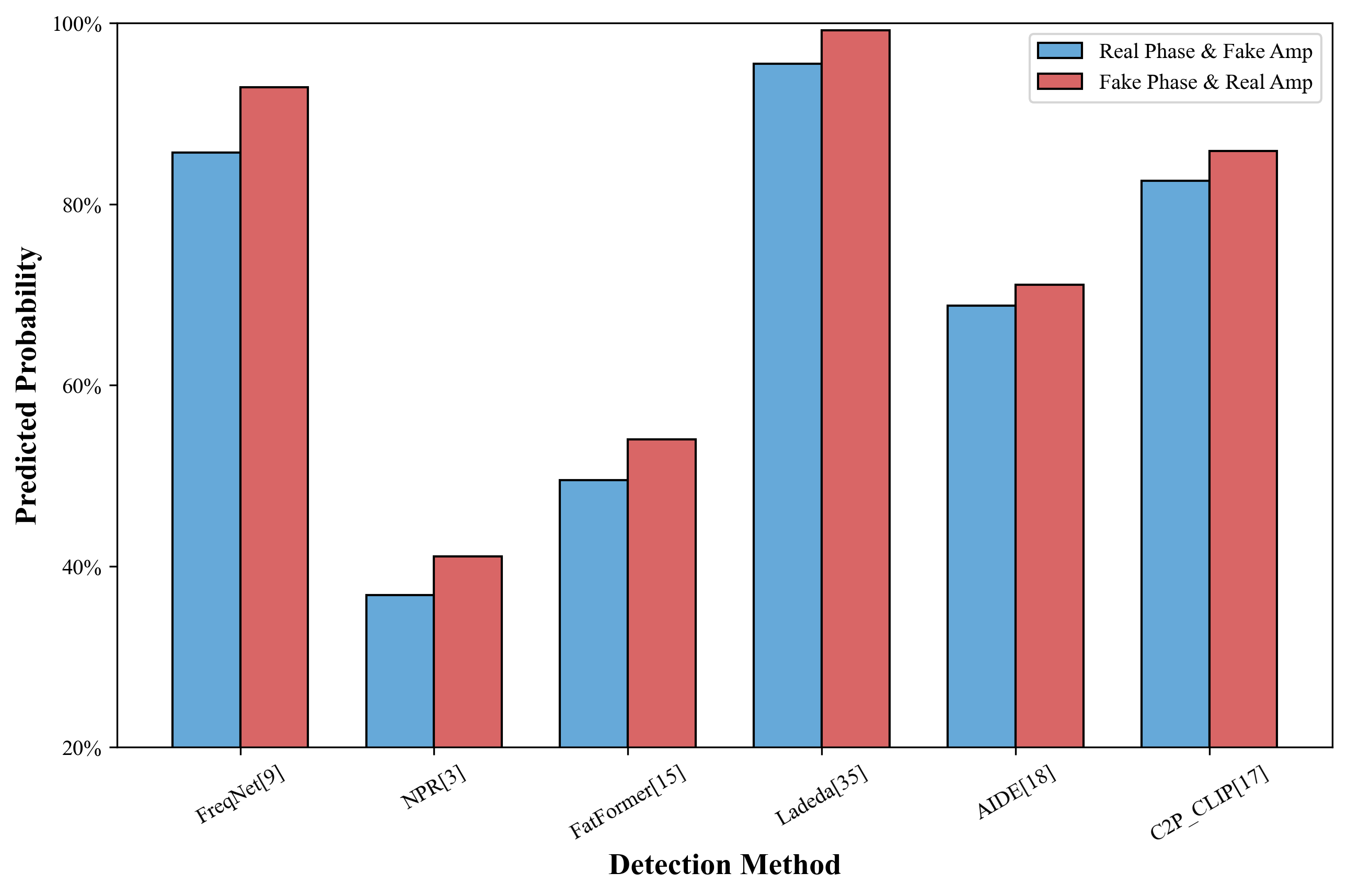}
    \caption{\textbf{Average Predicted Fake Probability in the Phase-Swapping Experiment.} A smaller value of the predicted fake probability indicates that the model is more likely to classify the image as real. The experimental results show that detectors generally consider images combining real phase and fake amplitude to be more realistic, while images combining fake phase and real amplitude are judged as fake, indicating that the phase component contains stronger forgery-related artifacts than the amplitude, where the amplitude mainly reflects color and brightness.}
    \label{fig:contract}
\end{figure}

To address above issues, it is urgent for the proposed method to tackle the following two difficulties. \textbf{1) How to model the importance difference and complex dependencies of elements within local regions?}  we adopt a sliding window to limit the attention mechanism to a local window and reconstruct the features within the window, making the attention mechanism more adaptively focused on pixel-level details while retaining the complex dependencies between pixels. Specifically, we concatenate features from the different frequency bands in each frequency domain. Then the nearby features are tiled in each frequency band. Finally, a sliding window is applied to extract the tiled features from all frequency bands. In this way, the reconstructed window features not only obtain features from the different frequency bands, but also retain the implicit relationships between adjacent elements. \textbf{2) How to extract more abundant and effective local features to further amplify the forgery traces?} It is well known that the DWT domain can reflect local spatial frequency features containing high-frequency and low-frequency texture information \cite{sadhya2024deepfake,badr2025wavit,liu2024forgery}. In this paper, we find that each element of the phase part in the FFT domain contains the dependencies and structural relationship between the overall elements. That means it can capture the longer-range dependencies between the global elements, thereby enriching the extracted fine-grained local features from another perspective. To verify the rationality of above finding, we randomly selected 1k real images and 1k fake images from the dataset GenImage, and cropped them to the same size. We randomly exchange the phase part of the real image and the fake image, and use the swapped images as the input of the detector. The average prediction results of the images predicted as fake are shown in Figure \ref{fig:contract}. The smaller the prediction result is, the more likely the image is to be classified as a real image. We find that the detector tends to judge images with fake phase as fake category rather than fake amplitude. This indicates that the phase part contains more artifact traces. Moreover, the amplitude part mainly contains properties such as color and brightness, which is difficult to reflect forgery traces, as described in \cite{zhao2024wavelet}. 

Therefore, we design a dual frequency branch framework consisting of four frequency bands in the DWT domain and the phase part in the FFT domain to extract richer local artifact features from multiple perspectives. Based on this, a reconstructed sliding window attention mechanism is designed to obtain more fine-grained and generalized local artifact features, thus improving the detection performance across various generation methods. To fully assess the generalizability of our method, we evaluate it on large image datasets generated by 65 distinct generation models\footnote{AttGAN, BEGAN, CramerGAN, InfoMaxGAN, MMDGAN, RelGAN, S3GAN, SNGAN, STGAN, ADM, DDPM, IDDPM, LDM, PNDM, VQDiffusion, SDv1, SDv2, Guided, Glide\_50\_27, Glide\_100\_10, Glide\_100\_27, LDM\_100, LDM\_200, LDM\_200\_cfg, DALLE, ProGAN, StyleGAN, StyleGAN2, BigGAN, CycleGAN, StarGAN, GauGAN, Deepfake, WFIR, ADM, Glide, Midjourney, SDv1.4, SDv1.5, VQDM, Wukong, DALLE2, R3GAN, StyleGAN3, StyleGAN-XL, StyleSwim, BlendFace, E4S, FaceSwap, InSwap, SimSwap, FLUX1-dev, Midjourney V6, GLIDE, DALLE-3, Imagen3, SD3, SDXL, BLIP, Infinite-ID, InstantID, IP-adapter, PhotoMaker, SocialRF, CommunityAI}.  Comprehensive experimental results show that our method achieves a significant improvement over state-of-the-art methods C2P-CLIP \cite{tan2025c2p} and AIDE \cite{yan2024sanity} by 5.8\% and 7.0\%. The overall contributions of the proposed method can be summarized as follows.

\begin{itemize}

\item We creatively reconstructed the sliding window attention mechanism, which effectively constructs the features within the window while limiting the attention within the local window, so that the attention mechanism can more accurately measure the importance and interdependence between internal elements in the local area.

\item We discovered that the phase part of FFT domain can reflect the artifact traces precisely in addition to the high and low frequency information of the DWT domain, and further designed the dual frequency branch framework to extract richer local forgery features from multiple perspectives.

\item Comprehensive experimental results have verified that our method exhibits strong generalization ability across 65 different generative methods. Compared with the state-of-the-art methods, the detection accuracy is improved by 2.13\%, which further highlights its superiority and universality.

\end{itemize}

\section{Related Work}

With the continuous development of generation methods, including GAN\cite{karras2017progressive,brock2018large,choi2018stargan,rossler2019faceforensics++}, diffusion \cite{dhariwal2021diffusion,rombach2022high,midjourney,FLUX_1}, customized generation \cite{wang2024instantid,wu2024infinite,li2024photomaker,li2022blip,ye2023ip} and so on, the generated images are becoming increasingly realistic, and the naked eye can no longer distinguish the authenticity of the images. In order to solve the future issues mentioned above, many researchers are committed to detecting AI-generated images. In this section, we will briefly introduce the existing detection methods.

\subsection{Deepfake Detection}
Early research on deepfake detection often focused on specific facial regions, such as the eyes and lips, to identify forgery traces \cite{haliassos2021lips,chen2022self}. However, as forgery techniques have advanced, these simple traces have increasingly been refined and obscured, rendering biometric-based detection methods insufficient for identifying sophisticated forgery images. To enhance cross-domain performance, a variety of universal detectors have been proposed based on frequency analysis \cite{qian2020thinking,frank2020leveraging,luo2021generalizing}, wavelet \cite{miao2023f}, forgery augmentations \cite{xia2024advancing,sun2024diffusionfake,javed2025enhancing}, ID information \cite{dong2023implicit,huang2023implicit}, and decoupled representation learning \cite{liang2022exploring,yan2023ucf,yan2024transcending,li2025critical}. Specifically, UCF \cite{yan2023ucf} extracts common forgery features across manipulation techniques; LSDA \cite{yan2024transcending} constructs and simulates variations within and across forgery features in the latent space, expanding the forgery feature space and enabling the learning of a more generalizable decision boundary. FPG \cite{xia2024advancing} is proposed to elevate the generalization of deepfake detectors from the perspective of real-time perception analysis, while Li \textit{et al.} propose a critical forgetting mechanism to force the detector to pay more attention to common subtle forgery traces that were previously overlooked, thereby enhancing the generalization capability.

\subsection{Exploration of Fine-Grained Features in AI-Generated Image Detection}
Experiments show that forgery traces are often embedded in fine details, which motivates people to extract local artifact information, including spatial and frequency domains. In terms of spatial domains, some research narrows the focus to local areas to extract artifact information, including limited receptive fields \cite{chai2020makes,cavia2024real}, cropping the image into smaller patches \cite{chen2021local,chen2020manipulated,zheng2024breaking,baraldi2025contrasting}, etc. Specifically, Chai \textit{et al.} \cite{chai2020makes} employ limited receptive fields to identify patches that render images detectable; Zheng \textit{et al.} \cite{zheng2024breaking} divide the image into multiple patches, and train a block-based convolutional network for feature extraction. In addition, LGrad \cite{tan2023learning} employs pretrained CNN models as transform functions to convert images into gradients and leverages these gradients to present universal artifacts. NPR \cite{tan2024rethinking} improves generalization by identifying upsampling artifacts, such as pixel grid distortions, in nearby regions. In the frequency domain, F3-Net \cite{qian2020thinking} introduces frequency component division and the frequency statistical difference between real images and forgery images into face forgery detection. FreDect \cite{frank2020leveraging} observes significant artifacts in the frequency domain of GAN-generated images, attributed to the upsampling operation in GAN architectures. Luo \textit{et al.} \cite{luo2021generalizing} utilize multiple high-frequency features of images to enhance generalization performance. FreqNet \cite{tan2024frequency} utilizes the Fast Fourier Transform to extract global high-frequency information, revealing refined forgery features such as periodic noise or anomalies invisible in the spatial domain. AIDE \cite{yan2024sanity} improved the generalization ability of forgery detection by extracting the highest and lowest fine-grained patches through DCT to capture low-level, fine-grained artifact information. What's more, some methods learn more generalizable artifact representations through large-scale data and augmentation strategies. For example, DRCT \cite{chen2024drct} leverages diffusion-based reconstruction to achieve better semantic alignment, while B-Free \cite{guillaro2025bias} mitigates dataset bias using self-conditioned inpainted reconstructions and content-aware augmentation.

\subsection{Application of Attention in AI-generated Image Detection}
Previous studies \cite{xu2015show,vaswani2017attention,dosovitskiy2020image} has proven that: The attention mechanism can not only be trained to focus on important areas of the image, but also capture the long-range dependencies of the image, which is very useful in AI-generated image detection. Many works have explored this: Zhao \textit{et al.}\cite{zhao2021multi} employ multiple spatial attention heads, allowing the model to attend to distinct regions—such as the eyes and mouth—simultaneously, enhancing its ability to spot inconsistencies. Wu \textit{et al.}\cite{wu2024local} introduce a local attention module interacting with distant structures (e.g., PPG graphs), though further details from the original study would clarify this mechanism. In recent years, the attention mechanism has been widely integrated into advanced models such as Vision Transformer (ViT) \cite{dosovitskiy2020image} and CLIP \cite{radford2021learning}, which not only improve the accuracy of detection, but also bring new possibilities for dealing with increasingly complex forgery techniques. For example, Ojha \textit{et al.} \cite{ojha2023towards} directly utilize image features from the CLIP model for linear classification; LASTED \cite{wu2023generalizable} proposes designing textual labels to supervise the CLIP vision model through image-text contrastive learning, further advancing the field of AIGC detection; Fatformer \cite{liu2024forgery} introduces a forgery-aware adapter to discern and integrate local forgery traces based on CLIP. CLIPMoLE \cite{liu2024mixture} adapts a combination of shared and separate LoRAs within an MoE-based structure in deeper ViT blocks. What's more, C2P-CLIP \cite{tan2025c2p} explored the reasons and potential of CLIP’s effectiveness in AI-generated image detection, and integrated category common cues into the text encoder to inject category-related concepts into the image encoder, thereby improving detection performance.

Existing attention mechanisms, such as Vision Transformers and CLIP, typically capture global weights and long-range dependencies to extract broad semantic information. However, they lack sensitivity to the significance and interrelationships of elements within local regions, hindering their ability to effectively detect fine-grained forgery traces. Therefore, we propose a reconstructed sliding window attention mechanism that enhances the feature extraction capacity of the window attention block to model the importance and dependencies of local forgery features. In addition, a dual frequency branch network is designed to enrich the features within the window from different angles, which promotes the extraction of more comprehensive local forgery features in the window.

\begin{figure*}
    \centering
    \includegraphics[width=1\linewidth]{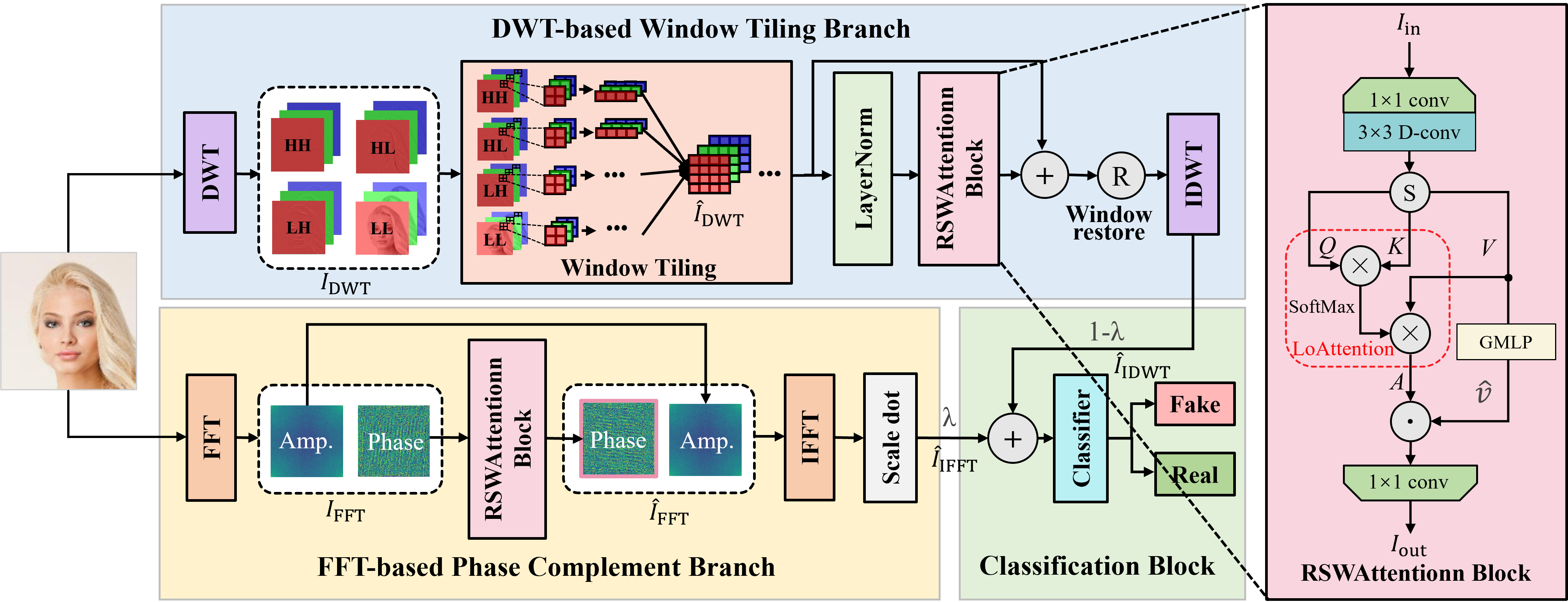}
    \caption{\textbf{Architecture of Our Method for Generalizable AI-Generated Image Detection}. 
    Specifically, we first design a reconstructed sliding window attention mechanism, which reconstructs the features within the sliding window and limits the attention mechanism to the local window range, forcing the attention to extract fine-grained forgery features while modeling the importance and dependencies between  internal elements in the local area. At the same time, we designed a dual frequency branch framework, which facilitates the model to extract richer artifact traces from multiple perspectives by using DWT-based window tiling and the phase part of FFT. The final extracted features are passed into a classifier.}
    \label{fig:enter-label}
\end{figure*}

\section{The Proposed Method}
\subsection{Main Backbone}
The overall structure of our method is shown in Figure \ref{fig:enter-label}. 
Specifically, we first propose a reconstructed sliding window attention mechanism, which reconstructs the features within the window while limiting the attention to the local window, forcing the attention to extract fine-grained forgery traces while modeling the importance and interdependence between internal elements within the local range. In addition, in order to enrich the extraction of local features from different perspectives, we design a dual frequency branch framework, using DWT-based window tiling and phase part of FFT before using window attention. In this way, the window feature not only obtains the features of different frequency bands, but also retains the implicit relationship between adjacent elements, while complementing the long-range dependency between features, which improves the overall generalization ability of the model.

\begin{figure}
    \centering
    \includegraphics[width=1\linewidth]{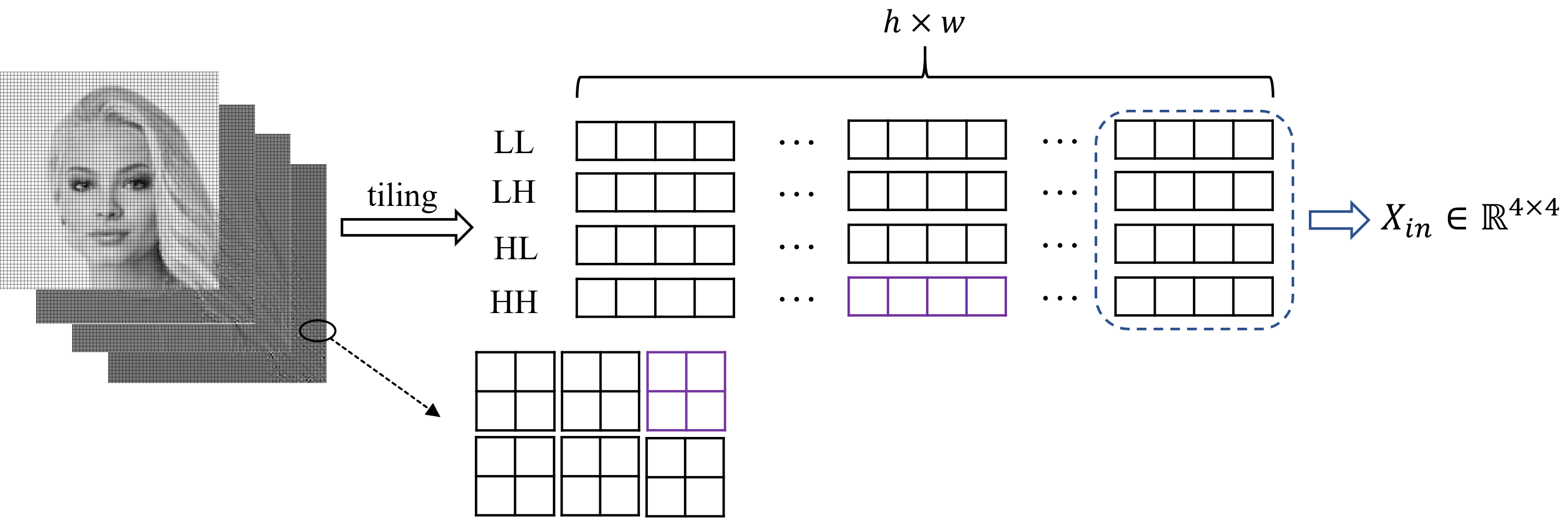}
    \caption{After DWT, sliding window tiling is used to tile the features of each channel into a shape of $ 4 \times (h \times w)$. In the subsequent LoAttention module, $4 \times 4$ features are extracted as shown in the blue box, which not only obtain features from the different frequency bands, but also retain the implicit relationships between adjacent elements. }
    \label{fig:split}
\end{figure}

\subsection{Reconstructed Sliding Window Attention Block}

The Reconstructed Sliding Window Attention Block (RSWAttention) consists of two main components: (i) the Local Attention Block (LoAttention) and (ii) the Global Window MLP Layer (GMLP). Before feeding the features into LoAttention and GMLP, we preprocess them to enhance their representational quality. A $1 \times 1$ convolution layer is applied to extract cross-channel features, followed by a $3 \times 3$ depthwise separable convolution to better capture the local features within each channel. The resulting convolved features are denoted as $I_{in}\in {\mathbb{R}}^{C \times H \times W}$.


\textbf{LoAttention.}  The attention mechanism obtains better results by training different weights of features. However, the weights of traditional attention mechanisms are derived from the features of the entire image, which contains a lot of semantic information. This weight design may cause the model to be disturbed when extracting fine-grained forgery traces. Thus we designed LoAttention module. Specifically, we employ a sliding window approach that divides the features into windows of size $ b \times b$. By restricting the attention module to these local windows, the focus shifts to effectively extracting local features. The features are then divided into $Q,K,V \in {\mathbb{R}}^{S \times C \times N}$, where $ S = \frac{H}{b}\times\frac{W}{b}$ and $N=b \times b $, as required by the attention module. Since the attention operates locally within each window, the features within window $i$ are represented as $Q_i,K_i,V_i \in {\mathbb{R}}^{C \times N}$. These localized features are fed into the attention module for computation: 
\begin{equation}
    A_i = \text{Attention}(Q_i, K_i, V_i) = \text{SoftMax}(Q_i K_i^{\top}) V_i.
\end{equation}

The results of all sliding window attention operations can be expressed as $A = \{A_1, A_2, \cdots, A_S\} \in {\mathbb{R}}^{S \times C \times N} $.

\textbf{GMLP.} To preserve global correlations while capturing fine-grained features, we integrate a global MLP within each window. This aims to address the limitations of local windows in modeling global relationships, ensuring a seamless fusion of local details and global context. Specifically, we reuse the value $V_i$ from LoAttention and process it through the global MLP, applying a nonlinear transformation to enhance its representational capacity. The result $\hat{V_i} $ is formally expressed as:
\begin{equation}
        \hat{V_i} \;=\; \text{GMLP}(V_i) \;=\; G(\text{Linear}(V_i)),
    \end{equation}
where $G(\cdot)$ denotes the GELU function. The GMLP recombines and redistributes the local window information, enabling the mixing of local frequency domain features and enhancing the learning of global features.

After combining the outputs from LoAttention and GMLP via a dot product, we use the inverse window operation to restore the feature to its original size $ C \times H \times W $, followed by a convolution to map the features back to the original feature dimensions. The final output of the RSWAttention $I_\text{out}$ can be expressed as:
\begin{equation}
    I_\text{out} = \text{conv}(\sum_{i=1}^{s}A_i \odot \sum_{i=1}^{s} \hat{V_i}).
\end{equation}

\subsection{Dual Frequency Branch Feature Enrichment Module}

\textbf{DWT-based window tiling branch.} Previous methods often leverage the Fast Fourier Transform (FFT) or Discrete Cosine Transform (DCT) to capture relationships between adjacent nodes in the frequency domain. However, these approaches can lead to the loss of positional information, which is crucial for extracting detailed features. To overcome this limitation, we utilize the Discrete Wavelet Transform (DWT), which decomposes the image into the spatial-frequency domain, represented as:
\begin{equation}
    \text{DWT}(I) = \{I_{LL}, I_{LH}, I_{HL}, I_{HH}\},
\end{equation}
where $I$ represents the input image, $I_{LL}, I_{LH}, I_{HL}, I_{HH}\in{\mathbb{R}^{C \times \frac{H}{2} \times \frac{W}{2}}} $ represents the low-frequency component of the input and the high-frequency components in the vertical, horizontal, and diagonal directions, respectively. To obtain more comprehensive information, we fuse the data from the four frequency domains, resulting in new input features $ I_\text{DWT} \in {\mathbb{R}^{4 \times C \times \frac{H}{2} \times \frac{W}{2}}} $. As illustrated in Figure \ref{fig:split}, to extract detailed features, we divide each channel’s features into $2 \times 2$ windows, with each window capturing localized spatial-frequency information within its respective frequency band. Each window is treated as a small square and tiled into rows within each band. This arrangement ensures that features within the $2\times2$ windows are preserved in a structured format. Then we tile the features in the window here by rows. At this time, the features from each frequency band are tiled into one row in a specific manner, such that the four adjacent feature values correspond to the $2\times2$ window from the original image. Additionally, as shown in Figure \ref{fig:split}, we integrate features from all frequency bands to generate new input features ${\hat{I}}_{\text{DWT}} \in {\mathbb{R}}^{C \times 4 \times \frac{H \times W}{4}}$. To further enhance the network’s performance, we apply LayerNorm \cite{lei2016layer}, which ensures that the input data remains within a stable range, reduces dependency on initialization, and improves the stability of model training.

The feature ${\hat{I}}_\text{DWT}$ is then fed into RSWAttention with a sliding window size set to $b=4$. As depicted in Figure \ref{fig:split}, within each $4 \times 4$ window, the representation includes not only the four adjacent feature values from the original image but also the feature information across four distinct frequency bands. This configuration provides RSWAttention with comprehensive and fine-grained spatial-frequency domain information, enabling more detailed feature extraction. Following this process, we apply the IDWT to reconstruct the features back into spatial domain representations, denoted as ${\hat{I}_{\text{IDWT}}}$.

\textbf{FFT-based phase complement branch.} Different from DWT, which captures local spatial frequency features, each element of the phase part in the FFT domain contains the dependencies and structural relationship between the overall elements, which helps the model capture the long-range dependencies between features. In addition, through our experimental analysis, we found that phase encodes finer details, including potential artifacts. Therefore, in order to enrich the extraction of local fine-grained features from another perspective, we only use the phase part of the FFT as another branch input of the RSWAttention, setting the window size to $b=8$. The amplitude remains unmodified during this process. After training the phase features, we use IFFT to recombine them with the original amplitude, transforming the features back into the spatial domain. The resulting features are denoted as ${\hat{I}_{\text{IFFT}}}$.

Finally, we set a hyperparameter $\lambda$ to effectively combines the outputs of two branches. The final feature is:
\begin{equation}
    {\hat{I}_{\text{output}}} = (1 - \lambda) *\hat{I}_{\text{IDWT}} + \lambda * \hat{I}_{\text{IFFT}}.
\end{equation}

\subsection{Classification Block}

After obtaining the fused representation $\hat{I}_{\text{output}}$ from the dual-frequency branches, we feed it into a classifier $f(\cdot)$ to produce the final logit. The entire network is trained end-to-end with a standard binary classification loss. Given a ground-truth label $y \in \{0,1\}$ (0: real, 1: fake), the objective is
\begin{equation}
    \mathcal{L}_{\mathrm{cls}} = -\, y \log \sigma(f(\hat{I}_{\text{output}})) - (1-y)\log (1-\sigma (f(\hat{I}_{\text{output}}))),
\end{equation}
where $\sigma(\cdot)$ denotes the sigmoid function. In our implementation, the classifier $f(\cdot)$ consists of a ResNet-50 backbone followed by a linear classification head.

\begin{table*}[ht!]
    \caption{Cross-model Accuracy (Acc.) \& Average Precision Score (A.P.) Performance on the GANGen-Detection Dataset \protect \cite{tan2024frequency}. }
    \label{tab_GANGen}
    \renewcommand\arraystretch{1.2}
    \resizebox{1.0\linewidth}{!}{
        \begin{tabular}{lcccccccccccccccccc|cc}
        \hline
        \multirow{2}{*}{Method} & \multicolumn{2}{c}{BEGAN} & \multicolumn{2}{c}{SNGAN} & \multicolumn{2}{c}{CramerGAN} & \multicolumn{2}{c}{MMDGAN} & \multicolumn{2}{c}{RelGAN} & \multicolumn{2}{c}{STGAN} & \multicolumn{2}{c}{S3GAN} & \multicolumn{2}{c}{AttGAN} & \multicolumn{2}{c}{InfoMaxGAN} & \multicolumn{2}{|c}{Mean} \\ 
        \cline{2-21} 
                                & Acc.         & A.P.       & Acc.         & A.P.       & Acc.          & A.P.          & Acc.         & A.P.        & Acc.          & A.P.       & Acc.          & A.P.      & Acc.         & A.P.       & Acc.          & A.P.       & Acc.             & A.P.        & Acc.       & A.P.        \\ 
        \hline
        CNN-Spot                & 50.2         & 44.9       & 62.7         & 90.4       & 81.5          & 97.5          & 72.9         & 94.4        & 53.3          & 82.1       & 63.0          & 92.7      & 55.2         & 66.1       & 51.1          & 83.7       & 71.1             & 94.7        & 62.3       & 82.9        \\
        Patchfor                & 97.1         & 100.0      & 97.6         & 99.8       & 97.8          & 99.9          & 97.9         & 100.0       & 99.6          & 100.0      & 92.7          & 99.8      & 66.8         & 68.8       & 68.0          & 92.9       & 93.6             & 98.6        & 90.1       & 95.4        \\
        F3Net                   & 87.1         & 97.5       & 51.6         & 93.6       & 89.5          & 99.8          & 73.7         & 99.6        & 98.8          & 100.0      & 60.3          & 99.9      & 65.4         & 70.0       & 85.2          & 94.8       & 67.1             & 83.1        & 75.4       & 93.1        \\
        GANDetection            & 67.9         & 100.0      & 66.7         & 90.6       & 67.8          & 99.7          & 67.7         & 99.3        & 60.9          & 86.2       & 69.6          & 97.2      & 69.6         & 83.5       & 57.4          & 75.1       & 67.6             & 92.4        & 66.1       & 91.6        \\
        LGrad                   & 69.9         & 89.2       & 78.0         & 87.4       & 50.3          & 50.3          & 57.5         & 67.3        & 89.1          & 99.1       & 54.8          & 68.0      & 78.5         & 86.0       & 68.6          & 93.8       & 71.1             & 82.0        & 68.6       & 80.8        \\
        UnivFD                 & 88.9         & 96.3       & 87.7         & 96.5       & 89.8          & 99.1          & 89.7         & 99.0        & 94.0          & 98.1       & 82.6          & 91.5      & 94.7         & 98.9       & 90.0          & 96.8       & 87.4             & 96.5        & 89.4       & 97.0        \\
        FreqNet                 & 98.8         & 100.0      & 85.4         & 90.4       & 95.2          & 98.2          & 95.2         & 98.2        & 100.0         & 100.0      & 98.8          & 100.0     & 88.3         & 94.3       & 89.8          & 98.8       & 94.0             & 97.5        & 94.0       & 97.5        \\
        NPR                     & 99.6         & 99.9       & 93.3         & 97.3       & 98.5          & 98.3          & 98.5         & 98.3        & 99.8          & 100.0      & 99.6          & 100.0     & 79.8         & 78.9       & 92.5          & 98.6       & 91.5             & 97.2        & 94.8       & 96.5        \\
        FatFormer               & 99.9         & 100.0      & 98.3         & 99.9       & 98.4          & 100.0         & 98.4         & 100.0       & 99.5          & 100.0      & 98.8          & 99.8      & 99.0         & 100.0      & 99.3          & 100.0      & 98.4             & 100.0       & \pmb{98.9}       & \pmb{100.0}       \\
        Ladeda                  & 100.0        & 100.0      & 99.0         & 99.9       & 97.9          & 99.9          & 99.0         & 99.9        & 100.0         & 100.0      & 99.8          & 100.0     & 80.5         & 84.9       & 100.0         & 100.0      & 98.4             & 99.9        & 97.2       & 98.3        \\
        C2P-CLIP                & 94.9         & 100.0      & 98.4         & 99.9       & 98.4          & 100.0         & 98.4         & 100.0       & 92.0          & 99.8       & 97.6          & 99.6      & 99.0         & 100.0      & 90.4          & 99.8       & 98.4             & 100.0       & 96.4       & \underline{99.9}        \\
        AIDE                    & 57.9         & 79.5       & 76.1         & 79.2       & 82.0          & 88.1          & 82.2         & 88.3        & 95.2          & 96.5       & 92.7          & 98.1      & 69.0         & 78.8       & 96.6          & 98.4       & 80.7             & 82.8        & 81.4       & 87.8        \\
        \rowcolor{mycolor}Ours  & 98.6         & 99.7       & 96.8         & 99.6       & 98.0          & 99.7          & 98.7         & 99.8        & 99.0          & 99.9       & 99.1          & 100.0     & 88.2         & 91.1       & 99.0          & 99.8       & 97.2             & 99.7        & \underline{97.3}       & 98.8        \\ 
        \hline
        \end{tabular}
    }
\end{table*}

\begin{table*}[ht!]
    \caption{Cross-model Accuracy (Acc.) \& Average Precision Score (A.P.) Performance on the DiffusionForensics Dataset\protect \cite{wang2023dire}.}
    \label{tab_Diffusion}
    \renewcommand\arraystretch{1.2}
    \resizebox{1.0\linewidth}{!}{
        \begin{tabular}{lcccccccccccccccc|cc}
        \hline
        \multirow{2}{*}{Method} & \multicolumn{2}{c}{DDPM} & \multicolumn{2}{c}{VQ-Diffusion} & \multicolumn{2}{c}{SDv1} & \multicolumn{2}{c}{ADM} & \multicolumn{2}{c}{SDv2} & \multicolumn{2}{c}{IDDPM} & \multicolumn{2}{c}{PNDM} & \multicolumn{2}{c}{LDM} & \multicolumn{2}{|c}{Mean} \\ \cline{2-19} 
                                & Acc.         & A.P.      & Acc.         & A.P.              & Acc.          & A.P.     & Acc.         & A.P.     & Acc.         & A.P.      & Acc.          & A.P.      & Acc.          & A.P.     & Acc.         & A.P.     & Acc.          & A.P.     \\      \hline
        CNN-Spot                & 62.7         & 76.6      & 50.0         & 71.0              & 38.0          & 76.7     & 53.9         & 71.8     & 52.0         & 90.3      & 50.2          & 82.7      & 50.8          & 90.3     & 50.4         & 78.7     & 51.0          & 90.3     \\
        Patchfor                & 62.3         & 97.1      & 100.0        & 100.0             & 90.7          & 99.8     & 77.5         & 93.9     & 94.8         & 100.0     & 50.0          & 91.6      & 50.2          & 99.9     & 99.5         & 100.0    & 78.1          & 97.8     \\
        F3Net                   & 84.7         & 99.4      & 100.0        & 100.0             & 73.4          & 97.2     & 80.9         & 96.9     & 99.8         & 100.0     & 74.7          & 98.9      & 72.8          & 99.5     & 100.0        & 100.0    & 85.8          & 99.0     \\
        GANDetection            & 62.3         & 46.4      & 51.1         & 51.2              & 39.8          & 65.6     & 51.1         & 53.1     & 50.1         & 36.9      & 50.2          & 63.0      & 50.6          & 79.0     & 51.6         & 48.1     & 50.8          & 55.4     \\
        LGrad                   & 99.9         & 100.0     & 96.2         & 100.0             & 90.4          & 99.4     & 86.4         & 97.5     & 97.1         & 100.0     & 66.1          & 92.8      & 69.5          & 98.5     & 99.7         & 100.0    & 88.2          & 98.5     \\
        UnivFD                 & 72.9         & 78.8      & 77.7         & 99.2              & 59.0          & 93.0     & 88.7         & 98.6     & 53.3         & 87.6      & 73.4          & 97.1      & 86.2          & 99.2     & 50.7         & 87.9     & 70.2          & 92.5     \\
        FreqNet                 & 91.4         & 99.8      & 100.0        & 100.0             & 63.9          & 98.1     & 67.2         & 91.3     & 81.8         & 98.4      & 59.0          & 97.3      & 85.2          & 99.8     & 98.9         & 100.0    & 80.9          & 98.1     \\
        NPR                     & 99.4         & 100.0     & 100.0        & 100.0             & 97.1          & 99.7     & 88.4         & 97.9     & 96.4         & 100.0     & 87.5          & 98.0      & 97.5          & 100.0    & 100.0        & 100.0    & \underline{95.8}          & 99.4     \\
        FatFormer               & 67.2         & 73.5      & 100.0        & 100.0             & 61.7          & 96.8     & 70.8         & 93.4     & 84.4         & 98.2      & 69.3          & 94.3      & 99.3          & 100.0    & 97.3         & 100.0    & 81.2          & 94.4     \\
        Ladeda                  & 98.4         & 99.9      & 100.0        & 100.0             & 97.1          & 99.8     & 83.9         & 96.9     & 94.1         & 100.0     & 93.9          & 98.8      & 93.7          & 99.9     & 100.0        & 100.0    & 95.1          & \underline{99.4}     \\
        C2P-CLIP                & 73.4         & 76.2      & 95.8         & 99.7              & 78.9          & 99.2     & 68.8         & 95.3     & 66.7         & 94.8      & 80.7          & 94.9      & 84.2          & 97.2     & 97.3         & 99.7     & 80.7          & 94.6     \\
        AIDE                    & 86.5         & 92.5      & 92.1         & 99.4              & 91.5          & 98.6     & 85.7         & 94.1     & 88.6         & 95.1      & 87.8          & 94.8      & 89.6          & 97.8     & 87.4         & 95.0     & 88.6          & 95.9     \\
        \rowcolor{mycolor}Ours  & 99.9         & 100.0     & 100.0        & 100.0             & 97.5          & 99.9     & 96.0         & 99.8     & 99.9         & 100.0     & 98.5          & 100.0     & 99.2          & 100.0    & 99.6         & 100.0    & \pmb{98.8}          & \pmb{100.0}    \\      \hline
        \end{tabular}
    }
\end{table*}

\section{Experiments}

To verify the excellent AI-generated image detection performance of our method, we follow the paradigms of \cite{ojha2023towards} and \cite{wang2020cnn}, which will be described in detail below.
\subsection{Dataset}

\textbf{Training set.} To ensure a consistent basis for comparison, we use the training set of ForenSynths\cite{wang2020cnn}. The training set consists of 20 different categories, each containing 18,000 synthetic images generated using ProGAN, and an equal number of real images from the LSUN dataset. Following the paradigms of \cite{ojha2023towards}, we adopt specific 4-class training settings, denoted as (car, cat, chair, horse).

\textbf{Testing set.} In order to comprehensively compare the effectiveness of our proposed method, we select different GAN and diffusion model-based generated datasets for testing, which contain a wide range of real-world images.
\begin{itemize}

\item GANGen-Detection \cite{tan2024frequency}: To evaluate more realistic scenarios, we extend our evaluation using 9 additional GAN-generated images. There are 4K test images for each model, with equal numbers of real and fake images.

\item DiffusionForensics \cite{wang2023dire}: To expand the testing scope, we adopt the diffusions dataset of DIRE \cite{wang2023dire} for evaluation, including ADM \cite{dhariwal2021diffusion}, DDPM \cite{ho2020denoising}, IDDPM \cite{nichol2021improved}, LDM \cite{rombach2022high}, PNDM \cite{liu2022pseudo}, VQ-Diffusion \cite{gu2022vector}, Stable Diffusion v1 \cite{rombach2022high}, Stable Diffusion v2 \cite{rombach2022high}. The real images are sampled from LSUN and ImageNet datasets.

\item UniversalFakeDetect \cite{ojha2023towards}: This test set contains images generated from Glide, DALLE \cite{ramesh2021zero}, LDM \cite{rombach2022high}. It adopts images of LAION and ImageNet datasets as real data.

\item AIGCDetectBenchmark \cite{zhu2024genimage}: This dataset comprehensively includes GAN and diffusion, and introduces some new diffusion generation methods as detection datasets, including ProGAN, StyleGAN, StyleGAN2, BigGAN, CycleGAN, StarGAN, GauGAN, Deepfake, WFIR, Midjourney, SDv1.4, SDv1.5, ADM, GLIDE, Wukong, VQDM, which is a huge challenge.

\item AIGIBench \cite{li2025artificial}: AIGIBench is the first benchmark to conduct a comprehensive evaluation of four critical components in the AIGI detection pipeline, and comprehensively simulates state-of-the-art image generation methods, including: (a) GAN-based noise-to-image generation (ProGAN, StyleGAN3, StyleGAN-XL, StyleSwim, R3GAN, and WFIR), (b) Diffusion for text-to-image generation (SD-XL, SD-3, DALLE-3, Midjourney-v6, FLUX.1-dev, Imagen-3, and GLIDE), (c) GANs for deepfake (BlendFace, E4S, FaceSwap, InSwap, and SimSwap), and (d) Diffusion for personalized generation (InstantID \cite{wang2024instantid}, Infinite-ID \cite{wu2024infinite}, PhotoMaker \cite{li2024photomaker}, BLIP-Diffusion \cite{li2022blip}, and IP-Adapter \cite{ye2023ip}). It also includes 2 general subsets collected from social media platforms, which called CommunityAI and SocialRF.

\end{itemize}

\subsection{Implementation Details}
We design a lightweight CNN network using convolutional layers and Resnet network as classifiers. The network is trained using the Adam optimizer\cite{kingma2014adam} with a learning rate of $2 \times 10^{-4}$. The batchsize is set to 128, and we train the network for 90 epochs. We also use a learning rate decay strategy that reduces the learning rate by twenty percent after every ten epochs. What's more, the hyperparameter $ \lambda $ is set to 0.4. For the loss function, we used BCELoss, which is commonly used in classification tasks. Our approach is implemented using PyTorch on an NVIDIA RTX A6000 GPU. To evaluate the performance of the proposed method, we follow the evaluation metrics used in baselines \cite{ojha2023towards,tan2024frequency}, which include average precision score (A.P.) and accuracy (Acc.).

\begin{table*}[ht!]
    \caption{Cross-model Accuracy (Acc.) \& Average Precision Score (A.P.) Performance on the UniversalFakeDetect Dataset \protect \cite{ojha2023towards}. }
    \label{tab_Universal}
    \renewcommand\arraystretch{1.2}
    \resizebox{1.0\linewidth}{!}{
        \begin{tabular}{lcccccccccccccccc|cc}
        \hline
        \multirow{2}{*}{Method} & \multicolumn{2}{c}{Guided} & \multicolumn{2}{c}{Glide\_50\_27} & \multicolumn{2}{c}{Glide\_100\_10} & \multicolumn{2}{c}{Glide\_100\_27} & \multicolumn{2}{c}{LDM\_100} & \multicolumn{2}{c}{LDM\_200} & \multicolumn{2}{c}{LDM\_200\_cfg} & \multicolumn{2}{c}{DALLE} & \multicolumn{2}{|c}{Mean} \\ \cline{2-19} 
                                & Acc.         & A.P.        & Acc.         & A.P.               & Acc.          & A.P.               & Acc.         & A.P.                & Acc.          & A.P.         & Acc.          & A.P.         & Acc.         & A.P.               & Acc.          & A.P.      & Acc.             & A.P.  \\         \hline
        CNN-Spot                & 54.9         & 66.6        & 54.2         & 76.0               & 53.3          & 72.9               & 53.0         & 71.3                & 51.9          & 63.7         & 52.0          & 64.5         & 51.6         & 63.1               & 51.8          & 61.3      & 52.8             & 67.4  \\
        Patchfor                & 74.2         & 81.4        & 84.9         & 98.8               & 87.3          & 99.7               & 82.8         & 99.1                & 95.8          & 99.8         & 95.6          & 99.9         & 94.0         & 99.8               & 79.8          & 99.1      & 86.8             & 97.2  \\
        F3Net                   & 69.2         & 70.8        & 88.5         & 95.4               & 88.3          & 95.4               & 87.0         & 94.5                & 74.1          & 84.0         & 73.4          & 83.3         & 80.7         & 89.1               & 71.6          & 79.9      & 79.1             & 86.5  \\
        GANDetection            & 50.1         & 51.0        & 51.7         & 53.5               & 51.2          & 52.6               & 51.1         & 51.9                & 54.7          & 65.8         & 54.9          & 65.9         & 53.8         & 58.9               & 67.2          & 83.0      & 54.3             & 60.1  \\
        LGrad                   & 71.8         & 76.0        & 91.6         & 96.0               & 90.9          & 95.5               & 88.5         & 94.2                & 95.7          & 99.3         & 94.7          & 99.1         & 95.1         & 99.0               & 89.9          & 97.8      & 89.8             & 94.6  \\
        UnivFD                 & 69.7         & 87.6        & 79.1         & 99.8               & 77.9          & 94.5               & 78.5         & 95.3                & 95.0          & 99.3         & 99.0          & 99.3         & 74.0         & 92.5               & 87.3          & 97.5      & 82.6             & 95.7  \\
        FreqNet                 & 67.3         & 75.7        & 86.7         & 96.3               & 87.9          & 96.4               & 84.5         & 96.0                & 97.9          & 99.9         & 97.5          & 99.9         & 97.4         & 99.9               & 97.4          & 99.8      & 89.6             & 95.5  \\
        NPR                     & 74.0         & 78.1        & 97.5         & 99.5               & 97.8          & 99.5               & 97.4         & 99.5                & 98.2          & 99.6         & 98.2          & 99.6         & 98.0         & 99.5               & 90.9          & 98.1      & 93.9             & 96.7  \\
        FatFormer               & 76.1         & 92.0        & 94.7         & 99.4               & 94.2          & 99.2               & 94.4         & 99.1                & 98.7          & 99.9         & 98.6          & 99.8         & 94.9         & 99.1               & 98.8          & 99.8      & 93.8             & 98.5  \\
        Ladeda                  & 79.8         & 87.5        & 98.3         & 99.8               & 98.2          & 99.8               & 98.5         & 99.8                & 98.6          & 99.9         & 98.5          & 99.9         & 98.0         & 99.8               & 83.9          & 98.5      & \underline{94.2}             & 98.1  \\
        C2P-CLIP                & 69.1         & 94.1        & 95.3         & 99.8               & 96.1          & 99.8               & 95.3         & 99.8                & 99.3          & 100.0        & 99.3          & 100.0        & 97.3         & 99.8               & 98.6          & 99.1      & 93.8             & \underline{99.2}  \\
        AIDE                    & 80.4         & 92.6        & 94.6         & 98.9               & 94.7          & 98.8               & 94.5         & 98.8                & 93.1          & 98.6         & 93.0          & 98.3         & 93.3         & 98.6               & 92.2          & 98.0      & 91.9             & 97.8  \\
        \rowcolor{mycolor}Ours  & 95.2         & 97.8        & 98.0         & 99.9               & 97.7          & 99.9               & 97.2         & 99.7                & 99.4          & 100.0        & 99.5          & 100.0        & 98.9         & 99.9               & 85.2          & 97.1      & \pmb{96.4}             & \pmb{99.3}  \\         \hline
        \end{tabular}
    }
\end{table*}

\begin{table*}[ht!]
    \caption{Cross-model Accuracy (Acc.) \& Average Precision Score (A.P.) Performance on the GenImage Datasets \protect \cite{zhu2024genimage}. }
    \label{tab_GenImage}
    \renewcommand\arraystretch{1.2}
    \resizebox{1.0\linewidth}{!}{
        \begin{tabular}{lcccccccccccccccccc}
        \toprule
        \multirow{2}{*}{Method} & \multicolumn{2}{c}{ProGAN} & \multicolumn{2}{c}{GauGAN} & \multicolumn{2}{c}{StyleGAN2} & \multicolumn{2}{c}{BigGAN} & \multicolumn{2}{c}{StyleGAN} & \multicolumn{2}{c}{CycleGAN} & \multicolumn{2}{c}{StarGAN} & \multicolumn{2}{c}{Deepfake} & \multicolumn{2}{c}{WFIR} \\ \cline{2-19} 
                                & Acc.         & A.P.        & Acc.         & A.P.        & Acc.          & A.P.          & Acc.         & A.P.        & Acc.          & A.P.         & Acc.          & A.P.         & Acc.         & A.P.          & Acc.         & A.P.         & Acc.             & A.P.  \\ \toprule
        CNN-Spot                & 97.9         & 99.9        & 66.1         & 93.5        & 67.0          & 97.1          & 55.9         & 80.5        & 70.1          & 97.0         & 76.5          & 91.5         & 72.9         & 93.2          & 51.7         & 66.5         & 62.1             & 89.4  \\
        LGrad                   & 99.9         & 100.0       & 72.4         & 79.3        & 96.0          & 99.9          & 82.9         & 90.7        & 94.8          & 99.9         & 85.3          & 94.0         & 99.6         & 100.0         & 58.0         & 67.9         & 60.5             & 65.1  \\
        UnivFD                 & 100.0        & 100.0       & 99.7         & 100.0       & 75.7          & 97.8          & 95.1         & 99.3        & 84.4          & 97.1         & 98.7          & 99.8         & 99.9         & 99.4          & 67.4         & 82.0         & 64.2             & 68.7  \\
        FreqNet                 & 99.6         & 100.0       & 93.4         & 98.6        & 88.0          & 99.5          & 90.5         & 96.0        & 90.2          & 99.7         & 95.8          & 99.6         & 85.7         & 99.8          & 88.9         & 94.4         & 48.0             & 49.6  \\
        NPR                     & 99.9         & 100.0       & 80.9         & 83.0        & 99.6          & 100.0         & 84.0         & 85.6        & 98.1          & 99.8         & 95.2          & 98.1         & 99.8         & 100.0         & 77.2         & 76.0         & 60.7             & 63.9  \\
        FatFormer               & 99.9         & 100.0       & 99.4         & 100.0       & 98.8          & 99.9          & 99.5         & 99.9        & 97.2          & 99.8         & 99.3          & 100.0        & 99.8         & 100.0         & 93.2         & 98.0         & 88.2             & 98.5  \\
        Ladeda                  & 100.0        & 100.0       & 94.1         & 100.0       & 99.9          & 100.0         & 82.5         & 89.5        & 99.8          & 100.0        & 91.1          & 99.0         & 76.2         & 90.3          & 64.0         & 92.6         & 85.7             & 93.6  \\
        C2P-CLIP                & 100.0        & 100.0       & 99.2         & 100.0       & 95.6          & 99.9          & 99.1         & 100.0       & 96.4          & 99.5         & 97.3          & 100.0        & 99.6         & 100.0         & 93.8         & 98.6         & 94.8             & 99.5  \\
        AIDE                    & 98.0         & 99.8        & 60.2         & 71.9        & 97.5          & 99.6          & 71.0         & 82.2        & 97.4          & 99.5         & 86.0          & 97.1         & 99.3         & 100.0         & 53.7         & 70.3         & 90.3             & 97.0  \\
        \rowcolor{mycolor}Ours  & 99.8         & 100.0       & 85.1         & 96.7        & 99.9          & 100.0         & 87.1         & 95.1        & 99.8          & 100.0        & 97.9          & 98.8         & 94.9         & 99.2          & 64.2         & 71.8         & 52.2             & 46.8  \\ \toprule
        \end{tabular}
    }
    \renewcommand\arraystretch{1.2}
    \resizebox{1.0\linewidth}{!}{
        \begin{tabular}{lcccccccccccccccc|cc}
        \hline
        \multirow{2}{*}{Method} & \multicolumn{2}{c}{ADM} & \multicolumn{2}{c}{Glide} & \multicolumn{2}{c}{Midjourney} & \multicolumn{2}{c}{SDv1.4} & \multicolumn{2}{c}{SDv1.5} & \multicolumn{2}{c}{VQDM} & \multicolumn{2}{c}{Wukong} & \multicolumn{2}{c}{DALLE2} & \multicolumn{2}{|c}{Mean} \\ \cline{2-19} 
                                & Acc.         & A.P.     & Acc.         & A.P.       & Acc.          & A.P.           & Acc.         & A.P.        & Acc.         & A.P.        & Acc.          & A.P.     & Acc.          & A.P.       & Acc.         & A.P.        & Acc.          & A.P.     \\         \hline
        CNN-Spot                & 52.1         & 67.8     & 51.7         & 67.1       & 51.7          & 64.7           & 50.6         & 59.2        & 50.6         & 59.2        & 51.5          & 63.5     & 50.5          & 58.5       & 50.1         & 49.2        & 61.3          & 76.7     \\
        LGrad                   & 65.8         & 70.2     & 60.6         & 73.8       & 64.6          & 70.3           & 65.9         & 68.7        & 66.3         & 69.1        & 67.2          & 70.6     & 63.2          & 65.0       & 66.9         & 86.9        & 65.1          & 71.8     \\
        UnivFD                 & 67.6         & 87.2     & 61.7         & 85.5       & 56.7          & 74.5           & 63.1         & 87.2        & 62.8         & 86.3        & 84.9          & 96.8     & 71.2          & 90.9       & 49.5         & 60.2        & 73.5          & 83.6     \\
        FreqNet                 & 83.3         & 91.5     & 81.7         & 88.9       & 69.9          & 79.0           & 64.3         & 74.3        & 64.9         & 75.5        & 81.7          & 89.6     & 57.7          & 67.0       & 55.1         & 54.6        & 69.8          & 77.6     \\
        NPR                     & 70.3         & 74.9     & 77.0         & 83.1       & 76.4          & 82.4           & 76.6         & 82.7        & 77.8         & 83.5        & 74.9          & 77.4     & 74.2          & 77.9       & 61.9         & 71.6        & 81.4          & 84.7     \\
        FatFormer               & 78.5         & 91.8     & 88.1         & 96.0       & 56.1          & 62.8           & 67.8         & 81.1        & 68.1         & 81.1        & 86.9          & 97.0     & 73.1          & 85.9       & 69.7         & 81.9        & 86.1          & 92.6     \\
        Ladeda                  & 74.0         & 84.4     & 90.6         & 96.6       & 86.9          & 93.5           & 89.8         & 95.5        & 90.6         & 95.8        & 76.6          & 84.5     & 87.6          & 93.0       & 83.0         & 93.9        & 86.6          & \underline{92.8}     \\
        C2P-CLIP                & 77.2         & 95.8     & 88.5         & 98.8       & 59.2          & 83.0           & 82.8         & 97.3        & 82.8         & 97.2        & 87.2          & 98.4     & 80.1          & 95.5       & 65.1         & 92.9        & \underline{88.2}          & 92.6     \\
        AIDE                    & 83.1         & 95.8     & 85.9         & 95.3       & 69.1          & 78.1           & 83.2         & 95.2        & 82.8         & 93.9        & 80.3          & 92.2     & 83.5          & 95.0       & 86.6         & 97.6        & 82.8          & 91.8     \\
        \rowcolor{mycolor}Ours  & 94.2         & 97.3     & 95.2         & 98.7       & 94.7          & 97.8           & 94.2         & 97.6        & 94.5         & 98.0        & 96.5          & 98.3     & 93.5          & 96.9       & 96.8         & 99.2        & \pmb{90.6}          & \pmb{93.6}     \\         \hline
        \end{tabular}
    }
\end{table*}

\subsection{Quantitative analysis}
We compare with the previous methods: CNN-Spot \cite{wang2020cnn}, Patchfor \cite{chai2020makes}, F3Net \cite{qian2020thinking}, FrePGAN \cite{jeong2022frepgan}, GANDetection \cite{mandelli2022detecting}, LGrad \cite{tan2023learning}, UnivFD \cite{ojha2023towards}, FreqNet \cite{tan2024frequency} ,NPR \cite{tan2024rethinking}, FatFormer \cite{liu2024forgery}, Ladeda \cite{cavia2024real}, C2P-CLIP \cite{tan2025c2p}, AIDE \cite{yan2024sanity}. We conduct comprehensive experiments on five datasets to verify the effectiveness of our method. Specifically, the results of Patchfor \cite{chai2020makes}, F3Net \cite{qian2020thinking}, FrePGAN \cite{jeong2022frepgan}, GANDetection \cite{mandelli2022detecting} all come from NPR \cite{tan2024rethinking},  and we obtain results of FatFormer \cite{liu2024forgery}, C2P-CLIP \cite{tan2025c2p} using the official pre-trained model. In addition, we retrain all other methods following the paradigms of \cite{ojha2023towards}, which adopts specific 4-class training settings, denoted as (car, cat, chair, horse).

\textbf{GANGen-Detection\protect \cite{tan2024frequency}. }Since our method uses ProGAN for training, we first perform cross-domain testing on GAN. As shown in Table \ref{tab_GANGen}, our detection on GAN has achieved good results. Specifically, it achieves mean Accuracy (Acc.) and mean Average Precision (A.P.) of 97.3\% and 98.8\%, respectively. Compared to Ladeda and C2P-CLIP, our method improves mean Acc. by 0.1\% and 0.9\%, respectively. However, on Fatformer, the accuracy is 1.1\% higher than our method, which needs to be improved.

\textbf{DiffusionForensics \cite{wang2023dire} \& UniversalFakeDetect \cite{ojha2023towards}. } To further evaluate the generalization ability of our method, we train on ProGAN and test on images generated by various diffusion models. The results are presented in Tables \ref{tab_Diffusion} and \ref{tab_Universal}. Despite being trained exclusively on ProGAN-generated images, our method demonstrates strong generalization capabilities across diverse diffusion models. Specifically, it achieves mean Accuracy (Acc.) and mean Average Precision (A.P.) of 98.8\% and 100.0\%, respectively. Compared with the best detection methods NPR and Ladeda, our method improves mean Acc. by 3.0\% and 3.7\%, respectively. However, methods Fatformer, C2P-CLIP, and AIDE, which have good results on other datasets, have an accuracy rate of less than 90\% on this dataset, and even C2P-CLIP has only 80.7\% accuracy. The most direct speculation for this result is that they used CLIP in the network skeleton, but the real reason is still worth further exploration. Additionally, on the UniversalFakeDetect dataset \cite{ojha2023towards}, our method achieves an average accuracy of 96.4\%, 2.2\% accuracy improvement over the best method.

\textbf{GenImage \protect \cite{zhu2024genimage}. }Compared with testing only on diffusion or GAN, GenImage not only includes the classic GAN generation method, but also adds new and more realistic diffusion generation datasets, such as DALLE2. The test results are shown in Table \ref{tab_GenImage}. Our method achieves a notable improvement in mean accuracy, outperforming C2P-CLIP and Ladeda by 2.4\% and 4.0\%, respectively, and reaching an accuracy of 90.6\%.

\textbf{AIGIBench \protect \cite{li2025artificial}. }AIGIBench proposes a more challenging forgery detection dataset to evaluate the performance of forgery detection models from different perspectives. The test results are shown in Table \ref{table_AIGIBench}, our detection has achieved good results. Specifically, it achieves mean Accuracy (Acc.) and mean Average Precision (A.P.) of 77.9\% and 82.0\%, respectively. Compared to Ladeda and C2P-CLIP, our method improves mean Acc. by 3.0\% and 5.9\%, respectively. Although our results are the best, there are still major problems and improvements: 1) Although our method has achieved an accuracy rate of more than 90\% in most GAN and diffusion scenarios, it still has an accuracy rate of only about 60\% in the dataset collected in unknown generated social scenarios. Other methods also have very low accuracy rates, which still needs further research and improvement. 2) When it comes to face-swapping forgery detection, the performance is generally low, even less than 50\%. Almost all methods have this problem, which is also a problem that needs to be improved.

\begin{table*}[ht!]
    \caption{Cross-model Accuracy (Acc.) \& Average Precision Score (A.P.) Performance on the AIGIBench Datasets \protect \cite{li2025artificial}.}
    \label{table_AIGIBench}
    \renewcommand\arraystretch{1.2}
    \resizebox{1.0\linewidth}{!}{
        \begin{tabular}{lcccccccccccccccccc}
        \toprule
        \multirow{2}{*}{Method} & \multicolumn{2}{c}{ProGAN} & \multicolumn{2}{c}{R3GAN}  & \multicolumn{2}{c}{StyleGAN3} & \multicolumn{2}{c}{StyleGAN-XL} & \multicolumn{2}{c}{StyleSwim} & \multicolumn{2}{c}{WFIR}     & \multicolumn{2}{c}{BlendFace} & \multicolumn{2}{c}{E4S} & \multicolumn{2}{c}{FaceSwap}  \\
        \cmidrule{2-19}
                                & Acc.         & A.P.        & Acc.         & A.P.        & Acc.          & A.P.          & Acc.         & A.P.             & Acc.          & A.P.          & Acc.          & A.P.         & Acc.         & A.P.            & Acc.         & A.P.    & Acc.         & A.P.           \\ \toprule
        CNN-Spot                & 97.8         & 100.0       & 49.9         & 48.4        & 64.0          & 91.6          & 49.7         & 60.0             & 58.0          & 88.9          & 62.1          & 89.4         & 49.3         & 34.2            & 49.6         & 35.2    & 50.0         & 44.3           \\
        LGrad                   & 98.0         & 99.9        & 65.6         & 69.1        & 59.9          & 64.3          & 61.2         & 62.8             & 68.1          & 75.0          & 55.6          & 53.1         & 38.6         & 39.4            & 39.2         & 39.4    & 28.2         & 33.3           \\
        UnivFD                 & 99.5         & 100.0       & 93.0         & 98.6        & 68.4          & 79.0          & 87.1         & 95.0             & 94.5          & 99.3          & 91.0          & 97.8         & 47.2         & 42.7            & 72.4         & 82.5    & 69.3         & 79.4           \\
        FreqNet                 & 99.3         & 100.0       & 62.3         & 56.8        & 83.0          & 92.4          & 79.8         & 84.1             & 80.8          & 91.8          & 58.5          & 48.9         & 23.3         & 34.1            & 25.8         & 34.7    & 40.4         & 43.4           \\
        NPR                     & 99.8         & 100.0       & 80.8         & 75.6        & 84.1          & 89.4          & 74.5         & 75.9             & 84.7          & 91.7          & 77.8          & 77.0         & 33.9         & 33.3            & 34.0         & 34.4    & 41.1         & 39.5           \\
        FatFormer               & 99.9         & 100.0       & 94.4         & 99.3        & 91.5          & 96.9          & 93.4         & 98.0             & 93.9          & 99.8          & 88.2          & 98.5         & 46.0         & 44.7            & 70.0         & 73.0    & 77.6         & 83.4           \\
        LaDeDa                  & 100.0        & 100.0       & 72.9         & 69.2        & 83.7          & 86.3          & 82.1         & 84.2             & 82.6          & 86.4          & 84.5          & 91.3         & 28.2         & 38.0            & 29.0         & 42.4    & 41.2         & 40.2           \\
        C2P-CLIP                & 100.0        & 100.0       & 95.4         & 99.8        & 93.8          & 98.5          & 94.1         & 98.1             & 95.2          & 100.0         & 94.8          & 99.5         & 55.2         & 60.5            & 69.8         & 76.1    & 73.5         & 82.2           \\
        AIDE                    & 98.0         & 99.8        & 89.0         & 94.8        & 79.2          & 77.8          & 84.4         & 87.7             & 85.2          & 85.1          & 90.3          & 97.0         & 43.4         & 43.8            & 39.4         & 38.3    & 46.7         & 43.7           \\
        \rowcolor{mycolor}Ours  & 99.8         & 100.0       & 94.6         & 94.9        & 93.2          & 98.2          & 82.3         & 86.4             & 94.7          & 98.6          & 52.2          & 46.8         & 44.1         & 44.0            & 45.1         & 46.4    & 50.8         & 66.8           \\ \toprule
        \end{tabular}
    }
    \vspace{3pt}
    \renewcommand\arraystretch{1.2}
    \resizebox{1.0\linewidth}{!}{
        \begin{tabular}{lcccccccccccccccccc}
        \hline
        \multirow{2}{*}{Method} & \multicolumn{2}{c}{InSwap} & \multicolumn{2}{c}{SimSwap}   & \multicolumn{2}{c}{FLUX1-dev}   & \multicolumn{2}{c}{Midjourney V6}  & \multicolumn{2}{c}{GLIDE}    & \multicolumn{2}{c}{DALLE-3}    & \multicolumn{2}{c}{Imagen3} & \multicolumn{2}{c}{SD3}    & \multicolumn{2}{c}{SDXL}\\
         \cline{2-19} 
                                & Acc.         & A.P.        & Acc.          & A.P.          & Acc.         & A.P.             & Acc.          & A.P.               & Acc.          & A.P.         & Acc.         & A.P.            & Acc.         & A.P.         & Acc.         & A.P.         & Acc.         & A.P.     \\         \hline
        CNN-Spot                & 50.1         & 44.3        & 49.7          & 45.3          & 50.3         & 50.3             & 49.9          & 39.3               & 51.5          & 67.3         & 50.4         & 53.8            & 51.0         & 50.3         & 52.8         & 66.3         & 52.6         & 65.5     \\
        LGrad                   & 37.7         & 38.5        & 36.1          & 38.1          & 64.6         & 70.0             & 59.2          & 62.1               & 56.7          & 66.2         & 49.5         & 48.6            & 61.4         & 66.9         & 66.2         & 71.4         & 64.0         & 66.3     \\
        UnivFD                 & 48.9         & 48.2        & 50.1          & 51.5          & 46.3         & 39.6             & 44.8          & 35.6               & 69.4          & 81.7         & 46.8         & 38.4            & 48.1         & 44.1         & 49.0         & 48.6         & 50.7         & 53.6     \\
        FreqNet                 & 37.5         & 42.1        & 36.5          & 41.9          & 78.5         & 87.3             & 57.6          & 57.7               & 75.8          & 77.4         & 66.2         & 61.0            & 73.6         & 80.7         & 77.6         & 84.0         & 86.4         & 93.6     \\
        NPR                     & 41.0         & 36.9        & 40.5          & 38.5          & 87.2         & 93.6             & 69.7          & 71.1               & 78.2          & 85.2         & 57.6         & 61.9            & 84.6         & 91.0         & 86.4         & 93.5         & 87.0         & 93.7     \\
        FatFormer               & 57.9         & 66.4        & 61.7          & 72.2          & 48.3         & 41.8             & 47.6          & 43.5               & 84.5          & 92.2         & 50.1         & 46.8            & 46.9         & 43.1         & 62.9         & 68.5         & 75.0         & 83.0     \\
        LaDeDa                  & 38.3         & 41.8        & 37.3          & 39.6          & 84.2         & 88.0             & 67.0          & 68.0               & 86.4          & 89.4         & 39.2         & 46.3            & 81.1         & 84.6         & 83.6         & 87.6         & 83.1         & 88.0     \\
        C2P-CLIP                & 59.5         & 69.0        & 67.9          & 78.5          & 47.6         & 46.8             & 51.2          & 50.2               & 85.1          & 93.4         & 64.4         & 75.4            & 45.2         & 46.5         & 53.5         & 65.6         & 59.3         & 76.0     \\
        AIDE                    & 46.1         & 43.0        & 47.2          & 46.8          & 84.9         & 86.6             & 77.9          & 78.1               & 90.0          & 89.3         & 56.9         & 57.0            & 86.4         & 90.8         & 89.7         & 94.1         & 88.9         & 90.0     \\
        \rowcolor{mycolor}Ours  & 52.1         & 63.7        & 48.4          & 57.7          & 93.2         & 96.8             & 86.4          & 92.7               & 94.2          & 98.4         & 46.2         & 49.0            & 93.6         & 96.0         & 95.4         & 97.6         & 95.9         & 98.7     \\         \hline
        \end{tabular}
    }
    \renewcommand\arraystretch{1.2}
    \resizebox{1.0\linewidth}{!}{
        \begin{tabular}{lcccccccccccccc|cc}
        \hline
        \multirow{2}{*}{Method} & \multicolumn{2}{c}{BLIP}      & \multicolumn{2}{c}{Infinite-ID}   & \multicolumn{2}{c}{InstantID} & \multicolumn{2}{c}{IP-Adapter}  & \multicolumn{2}{c}{PhotoMaker}     & \multicolumn{2}{c}{SocialRF} & \multicolumn{2}{c}{CommunityAI}     & \multicolumn{2}{|c}{Mean}    \\
         \cline{2-17} 
                                & Acc.         & A.P.           & Acc.         & A.P.               & Acc.          & A.P.          & Acc.         & A.P.             & Acc.          & A.P.               & Acc.          & A.P.         & Acc.         & A.P.                 & Acc.         & A.P.         \\ \hline
        CNN-Spot                & 49.8         & 44.7           & 50.8         & 55.8               & 50.1          & 67.1          & 50.0         & 50.1             & 49.6          & 51.5               & 50.3          & 46.3         & 49.9         & 50.5                 & 53.6         & 57.6         \\
        LGrad                   & 56.4         & 57.8           & 31.3         & 34.4               & 53.9          & 54.8          & 45.5         & 46.3             & 32.0          & 35.8               & 49.6          & 49.3         & 56.4         & 63.0                 & 53.4         & 56.2         \\
        UnivFD                 & 57.8         & 68.9           & 72.4         & 84.2               & 67.4          & 79.5          & 62.3         & 73.9             & 47.0          & 41.4               & 49.2          & 46.2         & 49.9         & 50.0                 & 63.3         & 66.4         \\
        FreqNet                 & 93.8         & 99.9           & 79.0         & 74.5               & 79.8          & 86.3          & 78.8         & 79.9             & 77.0          & 74.9               & 54.2          & 58.1         & 55.9         & 69.7                 & 60.2         & 61.7         \\
        NPR                     & 94.0         & 97.0           & 62.0         & 62.8               & 71.3          & 75.9          & 79.0         & 81.1             & 46.4          & 52.0               & 54.5          & 58.4         & 54.4         & 60.0                 & 68.2         & 70.8         \\
        FatFormer               & 82.8         & 92.3           & 83.3         & 91.2               & 69.3          & 77.4          & 67.1         & 74.1             & 46.4          & 37.4               & 56.9          & 65.1         & 51.9         & 69.9                 & 69.9         & 74.3         \\
        LaDeDa                  & 93.4         & 96.7           & 79.0         & 67.5               & 81.0          & 81.8          & 82.3         & 82.2             & 69.3          & 66.6               & 56.2          & 61.7         & 54.8         & 55.9                 & 68.8         & 71.3         \\
        C2P-CLIP                & 84.9         & 93.8           & 85.2         & 93.4               & 86.8          & 93.9          & 55.1         & 65.9             & 50.6          & 52.0               & 56.4          & 62.3         & 51.0         & 54.5                 & 71.0         & \underline{77.3}         \\
        AIDE                    & 92.7         & 96.7           & 88.9         & 87.9               & 87.6          & 89.0          & 87.1         & 91.0             & 83.4          & 80.9               & 56.8          & 62.1         & 51.3         & 52.8                 & \underline{74.9}         & 76.2         \\
        \rowcolor{mycolor}Ours  & 99.2         & 99.7           & 90.0         & 91.0               & 94.6          & 97.7          & 93.3         & 96.6             & 92.7          & 95.2               & 60.8          & 70.8         & 54.6         & 66.6                 & \pmb{77.9}         & \pmb{82.0}         \\         \hline
        \end{tabular}
    }
\end{table*}

\subsection{Qualitative Analysis}

\begin{figure*}[!t]
    \centering
    \includegraphics[width=1\linewidth]{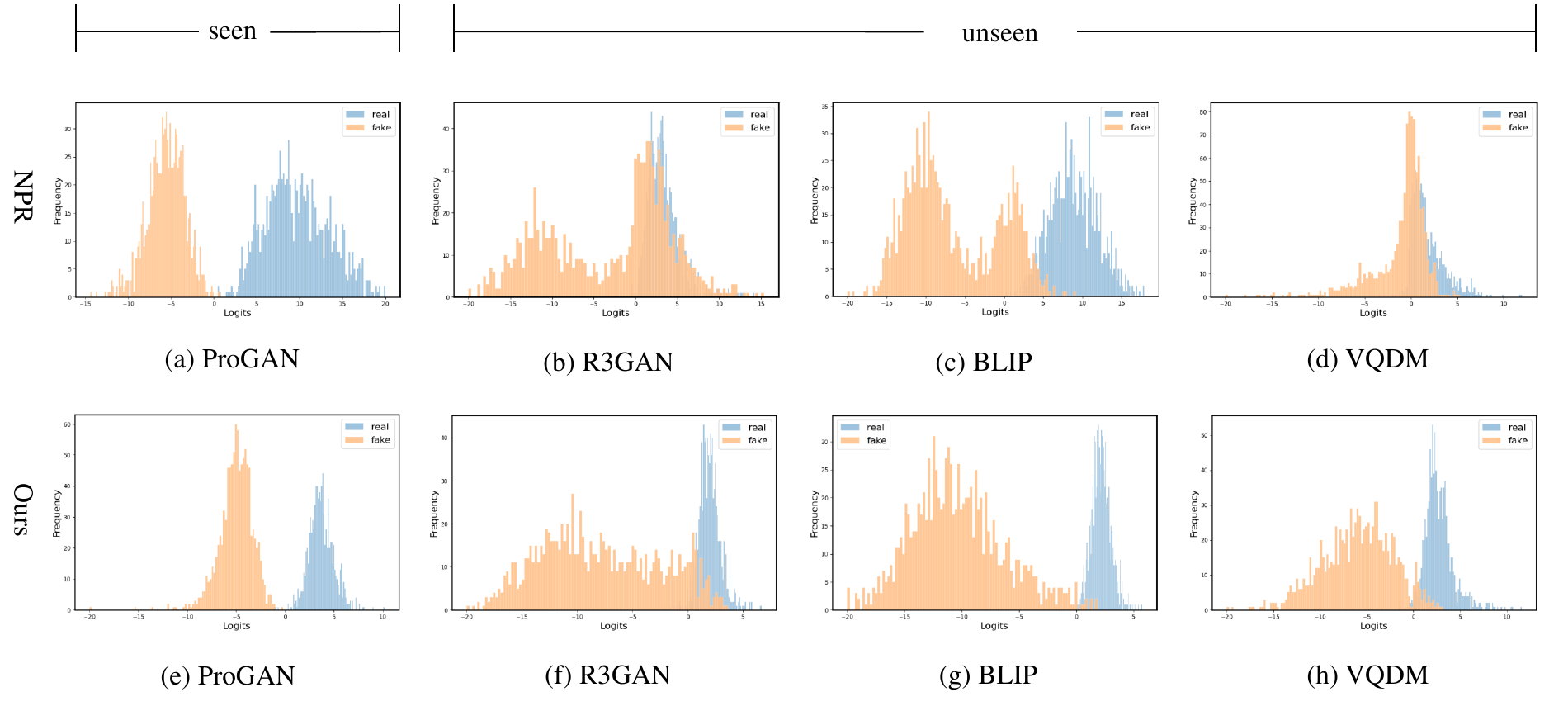}
    \caption{\textbf{Logit Distributions of Extracted Forgery Features}. We compare the state of the art NRP \protect \cite{tan2024rethinking} and our method, both tuned with 4-class ProGAN \protect \cite{karras2017progressive} data. A total of four testing GANs and diffusion models are considered, including ProGAN \protect \cite{karras2017progressive}, R3GAN \protect \cite{huang2024gan}, BLIP \protect \cite{li2022blip} and VQDM \protect \cite{gu2022vector}, each randomly sampled 1k real and 1k fake images.}
    \label{fig:logit}
\end{figure*}

To further assess the generalization ability of our method, we visualize the logit distributions of NPR and our approach, as shown in Figure \ref{fig:logit}. This visualization highlights how effectively each trained model distinguishes between real and fake images, showcasing our method's ability to generalize across diverse fake representations. From the visualization, it is evident that NPR struggles with unseen GAN or diffusion models, showing significant overlap between the logits of real and fake categories, often misclassifying fake images as "real." In contrast, our method demonstrates superior discrimination, effectively separating "real" and "fake" categories, even when encountering unseen sources.

\subsection{Feature Visualization of Dual Frequency Branches}
\begin{figure}[t]
    \centering
    \includegraphics[width=1\linewidth]{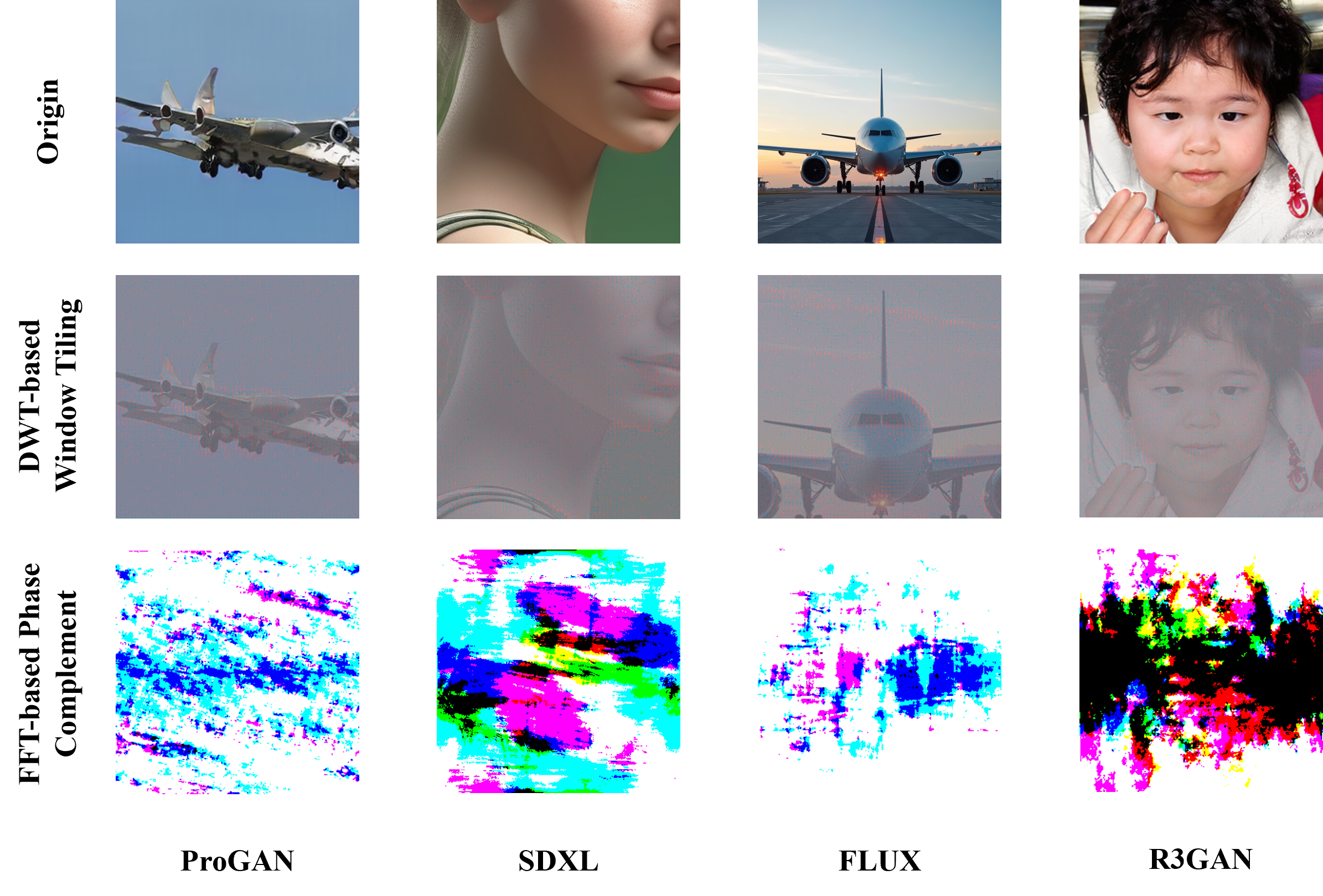}
    \caption{\textbf{Visualization after feature extraction from the two branches}. 
    The DWT-based Window Tiling mainly strengthens local edges and fine textures around object boundaries, whereas the FFT-based phase complement produces large-scale patch-like residual patterns, revealing non-local, generator-specific structural.}
    \label{fig:dwt_fft}
\end{figure}

To further validate that the two frequency branches capture complementary artifact information, we visualize the features extracted by each branch, as shown in Figure \ref{fig:dwt_fft}. Specifically, for the DWT-based window tiling branch, the reconstructed images exhibit intensified edges and subtle darkening around object contours, indicating that this branch focuses on fine-grained local artifacts and high-frequency inconsistencies. In contrast, the FFT-based phase complement branch yields patch-like residual patterns that spread over large regions rather than aligning with specific edges, which arise from phase perturbations in the frequency domain and reflect global structural irregularities that are characteristic of different generators. Together, these observations support the effectiveness of our method in extracting artifact cues from complementary perspectives.

\begin{table*}[]
    \caption{Module Ablation Studies on the Five Datasets. }
    \label{tab_ablation}
    \renewcommand\arraystretch{1.8}
    \resizebox{1.0\linewidth}{!}{
        \begin{tabular}{cccc|ccccc|c}
        \hline
        DWT                             & Window Tiling & FFT       & RSWAttention   & GANGen-Detection      & DiffusionForensics    & UniversalFakeDetect       & GenImage          & AIGIBench         & mean Acc. / A.P.      \\ \hline
        \xmark                          &\xmark         &\cmark     &\cmark          & 49.6 / 49.6           & 48.6 / 48.6           & 47.9 / 47.9               & 49.4 / 49.4       & 49.6 / 49.6       & 49.0 / 49.0           \\
        \cmark                          &\xmark         &\xmark     &\cmark          & 93.2 / 96.5           & 95.6 / 100.0          & 93.2 / 95.3               & 87.5 / 91.2       & 74.1 / 78.5       & 88.7 / 91.2           \\
        \cmark                          &\cmark         &\xmark     &\cmark          & 95.2 / 97.1           & 96.9 / 100.0          & 95.0 / 97.4               & 88.7 / 92.5       & 75.5 / 79.8       & 90.3 / \underline{93.4}           \\
        \cmark                          &\xmark         &\cmark     &\cmark          & 96.2 / 98.8           & 96.6 / 99.9           & 96.2 / 99.1               & 90.0 / 92.1       & 74.2 / 75.8       & \underline{90.6} / 93.1           \\ \hline
        \rowcolor{mycolor}\cmark        &\cmark         &\cmark     &\cmark          & \pmb{96.6 / 98.8}           & \pmb{98.8 / 100.0}          & \pmb{96.4 / 99.3}               & \pmb{90.6 / 93.6}       & \pmb{77.9 / 82.0}       & \pmb{92.1 / 94.1}           \\ \hline
        \end{tabular}
    }
\end{table*}

\begin{table*}[t]
    \centering
    \begin{minipage}[t]{0.48\linewidth}
        \centering
        \caption{Robustness on JPEG Compression and Gaussian Blur of Our Method. The accuracy (\%) averaged on GenImage \cite{zhu2024genimage}.}
        \label{table:Robustness_GenImage}
        \resizebox{\linewidth}{!}{
        \renewcommand{\arraystretch}{1.4}
        \begin{tabular}{l|ccc|ccc}
        \hline
        \multirow{2}{*}{Method} & \multicolumn{3}{c|}{JPEG Compression} & \multicolumn{3}{c}{Gaussian Blur} \\
                                & QF=95         & QF=85         & QF=75             & $\sigma = 0.5$        & $\sigma = 1.0$        & $\sigma = 1.5$ \\ \hline
    CNN-Spot                    & 59.0          & 59.9          & 60.5              & 62.4                  & 63.2                  & 63.3                  \\
    UnivFD                      & \underline{76.4}          & \underline{74.2}          & \underline{73.8}              & 80.2                  & 76.8                  & 76.3                  \\
    NPR                         & 72.8          & 70.7          & 69.2              & 80.6                  & 79.6                  & 81.0                  \\
    LaDeDa                      & 74.4          & 70.4          & 68.0              & 84.8                  & \underline{83.2}                  & \underline{83.5}                  \\
    AIDE                        & 74.6          & 74.1          & 70.3              & \underline{88.6}                  & 75.8                  & 81.1                  \\ \hline
    \rowcolor{mycolor}Ours      & \textbf{78.8}          & \textbf{76.6}          & \textbf{75.2}              & \textbf{92.0}                  & \textbf{88.4}                  & \textbf{87.9}                  \\ \hline
        \end{tabular}
        }
    \end{minipage}
    \hfill
    \begin{minipage}[t]{0.48\linewidth}
        \centering
        \caption{Robustness on JPEG Compression and Gaussian Blur of Our Method. The accuracy (\%) averaged on AIGIBench \cite{li2025artificial}.}
        \label{table:Robustness_AIGIBench}
        \resizebox{\linewidth}{!}{
        \renewcommand{\arraystretch}{1.4}
        \begin{tabular}{l|ccc|ccc}
        \hline
        \multirow{2}{*}{Method} & \multicolumn{3}{c|}{JPEG Compression} & \multicolumn{3}{c}{Gaussian Blur} \\
                                & QF=95         & QF=85         & QF=75             & $\sigma = 0.5$        & $\sigma = 1.0$        & $\sigma = 1.5$ \\ \hline
        CNN-Spot                    & 53.8          & 54.3          & 54.5              & 56.5                  & 56.5                  & 56.6                  \\
        UnivFD                      & \underline{65.6}          & \underline{64.3}          & \underline{62.7}              & 72.2                  & \underline{68.6}                  & \underline{68.6}                  \\
        NPR                         & 59.0          & 58.9          & 58.6              & 62.9                  & 62.0                  & 62.9                  \\
        LaDeDa                      & 61.6          & 60.0          & 58.2              & 64.4                  & 66.0                  & 65.6                  \\
        AIDE                        & 61.0          & 59.6          & 55.9              & \underline{75.5}                  & 63.4                  & 65.2                  \\ \hline
        \rowcolor{mycolor}Ours      & \textbf{68.9}          & \textbf{67.2}          & \textbf{65.2}              & \textbf{75.5}                  & \textbf{74.6}                  & \textbf{74.2}                  \\ \hline
        \end{tabular}
        }
    \end{minipage}
\end{table*}

\begin{table*}[]
    \caption{RSWAttention Validity Studies on the Five Datasets. }
    \label{tab_RSWAttention}
    \renewcommand\arraystretch{1.6}
    \resizebox{1.0\linewidth}{!}{
        \begin{tabular}{cc|ccccc|c}
        \hline
        & Attention                         & GANGen-Detection      & DiffusionForensics    & UniversalFakeDetect       & GenImage          & AIGIBench         & mean Acc. / A.P.      \\ \hline
        & WAttention                        & 70.7 / 82.9           & 65.8 / 89.5           & 73.2 / 88.4               & 71.5 / 81.2       & 58.3 / 63.2       & 67.9 / 81.0           \\
        \rowcolor{mycolor}& RSWAttention    & \pmb{96.6 / 98.8}           & \pmb{98.8 / 100.0}          & \pmb{96.4 / 99.3}               & \pmb{90.6 / 93.6}       & \pmb{77.9 / 82.0}       & \pmb{92.1 / 94.1}           \\ \hline
        \end{tabular}
    }
\end{table*}

\begin{table*}[]
    \caption{Phase-only Part of FFT Validity Studies on the Five Datasets. }
    \label{tab_fft}
    \renewcommand\arraystretch{1.6}
    \belowrulesep=0pt
    \aboverulesep=0pt
    \resizebox{1.0\linewidth}{!}{
        \begin{tabular}{cc|ccccc|c}
        \hline
        \multicolumn{2}{c|}{FFT}            & \multirow{2}{*}{GANGen-Detection}     & \multirow{2}{*}{DiffusionForensics}    & \multirow{2}{*}{UniversalFakeDetect}       & \multirow{2}{*}{GenImage}          & \multirow{2}{*}{AIGIBench}         & \multirow{2}{*}{mean Acc. / A.P.}      \\ \cmidrule{1-2}
        Phase      &Amp                     &                                       &                                        &                                            &                                    &                                    &                                        \\ \hline        
        \xmark     & \xmark                 & 95.2 / 97.1                           & 96.9 / 100.0                           & 95.0 / 97.4                                & 88.7 / 92.5                        & 75.5 / 79.8                        & \underline{90.3} / \underline{93.4}                \\
        \rowcolor{mycolor} \cmark &\xmark   & \pmb{96.6 / 98.8}                     & \pmb{98.8 / 100.0}                     & \pmb{96.4 / 99.3}                          & \pmb{90.6 / 93.6}                  & \pmb{77.9 / 82.0}                  & \pmb{92.1 / 94.1}                      \\
        \cmark     & \cmark                 & 93.0 / 97.0                           & 91.3 / 99.2                            & 92.2 / 97.2                                & 85.1 / 91.2                        & 72.7 / 77.8                        & 86.9 / 92.5                            \\ \hline
        \end{tabular}
    }
\end{table*}

\subsection{Module Ablation Studies} 
In order to comprehensively evaluate the effectiveness of our method, we conduct ablation studies on each module. The results of these experiments, shown in Table \ref{tab_ablation}, demonstrate the contribution of each module to AI-generated image detection. Specifically, the experiment can prove that: 

\textbf{FFT alone has negative impacts on Attention.} When we remove our main module (DWT-based Window Tiling), the performance of our method drops significantly, falling below 50\%. In our design, FFT is integrated as a parallel branch to complement DWT, leveraging its phase information to capture long-range dependencies that enhance fine-grained feature richness for reconstructed sliding window attention. However, the localized focus of RSWAttention, optimized for fine-grained details, appears incompatible with FFT’s global frequency representation when not moderated by DWT’s multi-scale decomposition or Window Tiling’s spatial structuring. This misalignment suggests that FFT alone introduces overly coarse features, diluting the attention mechanism’s ability to discern subtle, localized artifacts critical for detection.

\textbf{Reasonable design of sliding window greatly improves the ability of attention extraction artifacts.} Whether it is DWT or window tiling, the lack of any part will lead to a decrease in model accuracy. Without DWT, window attention cannot capture fine-grained artifacts. Removing Window Tiling yields a mean Acc. / A.P. of 90.3\% / 93.4\%, a notable decline from the full model’s 92.1\% / 94.1\%, suggesting that Window Tiling preserves critical spatial relationships within local windows, allowing RSWAttention to exploit Dependencies between local features. Without it, the disruption of these relationships impairs the model’s ability to detect localized artifacts, which is very important in datasets with complex or subtle manipulations. Therefore, carefully designed in-window features based on DWT and window tiling are crucial for RSWAttention to extract fine-grained forgery features. 

\textbf{It is effective to use the phase part of FFT as another branch.} As shown in Table \ref{tab_ablation}, when the phase part of FFT is excluded, the accuracy of model decreases 1.8\%, indicating that FFT enhances the attention mechanism’s capacity to model interdependencies between distant image regions. FFT, as another frequency domain branch, enriches the local features within the window attention from different perspectives. 

\begin{table*}[!ht]
    \caption{Comparative Studies Using FFT and DCT on the Five Datasets. }
    \label{tab_dct}
    \renewcommand\arraystretch{1.6}
    \resizebox{1.0\linewidth}{!}{
        \begin{tabular}{cc|ccccc|c}
        \hline
        & Complement Branch             & GANGen-Detection      & DiffusionForensics    & UniversalFakeDetect       & GenImage          & AIGIBench         & mean Acc. / A.P.      \\ \hline
        & DCT                           & 95.9 / 98.1           & 97.2 / 100.0          & 95.8 / 99.0               & 87.9 / 93.3       & 75.7 / 80.1       & 90.5 / 92.5           \\
        \rowcolor{mycolor}& FFT         & \pmb{96.6 / 98.8}           & \pmb{98.8 / 100.0}          & \pmb{96.4 / 99.3}               & \pmb{90.6 / 93.6}       & \pmb{77.9 / 82.0}       & \pmb{92.1 / 94.1}           \\ \hline
        \end{tabular}
    }
\end{table*}

\begin{table*}[]
    \caption{DWT Different Frequency Domain Subbands Studies on the Five Datasets. }
    \label{tab_dwt}
    \renewcommand\arraystretch{1.6}
    \resizebox{1.0\linewidth}{!}{
        \begin{tabular}{c|ccccc|c}
        \hline
        DWT                         & GANGen-Detection      & DiffusionForensics    & UniversalFakeDetect       & GenImage          & AIGIBench         & mean Acc. / A.P.      \\ \hline
        LL                          & 65.7 / 79.9           & 63.6 / 84.7           & 68.5 / 82.2               & 66.0 / 78.0       & 57.3 / 62.3       & 64.2 / 77.4           \\
        LH                          & 75.9 / 89.3           & 68.3 / 87.0           & 74.3 / 89.1               & 70.4 / 82.3       & 60.2 / 66.0       & 69.8 / 82.7           \\
        HL                          & 78.9 / 88.3           & 69.7 / 86.2           & 74.5 / 85.6               & 71.8 / 82.7       & 61.7 / 66.2       & 71.3 / 81.8           \\
        HH                          & 77.3 / 88.7           & 67.2 / 87.2           & 74.6 / 85.9               & 71.9 / 84.6       & 61.5 / 67.7       & 70.5 / 82.8           \\
        HL,HH                       & 91.8 / 96.3           & 81.4 / 99.3           & 91.6 / 97.7               & 81.4 / 90.3       & 71.5 / 76.7       & 80.3 / 88.0           \\
        LH,HL,HH                    & 90.7 / 98.0           & 84.0 / 99.9           & 92.5 / 99.0               & 84.3 / 93.0       & 73.8 / 80.5       & \underline{85.1} / \underline{94.0}           \\ 
        \rowcolor{mycolor}LL,LH,HL,HH       & \pmb{96.6 / 98.8}           & \pmb{98.8 / 100.0}          & \pmb{96.4 / 99.3}               & \pmb{90.6 / 93.6}       & \pmb{77.9 / 82.0}       & \pmb{92.1 / 94.1}           \\ \hline
        \end{tabular}
    }
\end{table*}

\begin{table*}[]
    \caption{Different Window Size Settings Studies on the Five Datasets. }
    \label{tab_window}
    \renewcommand\arraystretch{1.6}
    \resizebox{1.0\linewidth}{!}{
        \begin{tabular}{ccc|ccccc|c}
        \hline
        Window Tiling                   & DWT-Based     & FFT-Based     & GANGen-Detection      & DiffusionForensics    & UniversalFakeDetect       & GenImage          & AIGIBench         & mean Acc. / A.P.      \\ \hline
        $2 \times 2$                    & $4 \times 4$  & $4 \times 4$  & 92.5 / 98.2           & 83.8 / 99.8           & 93.8 / 99.0               & 87.4 / 93.1       & 71.4 / 77.1       & 85.4 / \underline{92.3}           \\
        $4 \times 4$                    & $4 \times 4$  & $8 \times 8$  & 95.8 / 98.8           & 91.1 / 99.9           & 95.9 / 99.3               & 89.2 / 92.2       & 73.8 / 76.7       & \underline{88.3} / 92.0           \\
        $4 \times 4$                    & $16 \times 16$& $8 \times 8$  & 92.9 / 97.0           & 84.0 / 99.2           & 89.4 / 97.7               & 82.1 / 91.0       & 64.6 / 72.1       & 81.1 / 89.5           \\
        $2 \times 2$                    & $2 \times 2$  & $8 \times 8$  & 74.4 / 80.7           & 70.4 / 89.7           & 66.4 / 73.1               & 67.4 / 72.6       & 59.6 / 66.2       & 68.6 / 76.4           \\
        $2 \times 2$                    & $8 \times 8$  & $8 \times 8$  & 95.5 / 98.3           & 91.9 / 99.9           & 94.8 / 99.0               & 88.9 / 91.9       & 71.8 / 74.7       & 87.4 / 91.3           \\ \hline
        \rowcolor{mycolor}$2 \times 2$  & $4 \times 4$  & $8 \times 8$  & \pmb{96.6 / 98.8}           & \pmb{98.8 / 100.0}          & \pmb{96.4 / 99.3}               & \pmb{90.6 / 93.6}       & \pmb{77.9 / 82.0}       & \pmb{92.1 / 94.1}           \\ \hline
        \end{tabular}
    }
\end{table*}

\begin{table*}[]
    \caption{Hyperparameter $\lambda$ Ablation Experiment on the Five Datasets. }
    \label{tab_lambda}
    \renewcommand\arraystretch{1.4}
    \resizebox{1.0\linewidth}{!}{
        \begin{tabular}{c|ccccc|cc}
        \hline
        $\lambda$                 & GANGen-Detection      & DiffusionForensics    & UniversalFakeDetect       & GenImage          & AIGIBench      & mean Acc. & mean A.P. \\ \hline
        0.0                     & 95.2 / 97.1           & 96.9 / 100.0          & 95.0 / 97.4               & 88.7 / 92.5       & 75.5 / 79.8    & 90.3      & 93.4      \\
        0.1                     & 95.7 / 98.9           & 97.2 / 100.0          & 95.8 / 99.0               & 89.4 / 93.0       & 77.0 / 81.1    & 91.0      & 94.4      \\
        0.2                     & 96.2 / 98.8           & 97.5 / 100.0          & 95.3 / 98.9               & 90.0 / 93.2       & 77.0 / 81.6    & \underline{91.2}      & \underline{94.5}      \\
        \rowcolor{mycolor}\pmb{0.4}   & \pmb{96.6 / 98.8}           & \pmb{98.8 / 100.0}          & \pmb{96.4 / 99.3}               & \pmb{90.6 / 93.6}       & \pmb{77.9 / 82.0}    & \pmb{92.1}      & \pmb{94.7}      \\ 
        0.6                     & 95.5 / 98.9           & 98.5 / 99.9           & 96.0 / 99.0               & 89.9 / 93.1       & 76.9 / 80.9    & 91.1      & 94.4      \\
        0.8                     & 95.0 / 97.8           & 93.4 / 100.0          & 93.7 / 98.5               & 89.9 / 94.2       & 76.5 / 80.2    & 89.7      & 94.1      \\
        1.0                     & 49.6 / 49.6           & 48.6 / 48.6           & 47.9 / 47.9               & 49.4 / 49.4       & 49.6 / 49.6    & 49.0      & 49.0      \\ \hline
        \end{tabular}
    }
\end{table*}

\subsection{Robustness Evaluation}
In real-world scenarios, images are inevitably affected by unknown disturbances during transmission and interaction, which pose additional challenges for AI-generated image detection. To investigate robustness under such conditions, we further evaluate various detection methods against common perturbations, including JPEG compression (Quality Factor (QF) = 95, 85, 75) and Gaussian blur ($\sigma$ = 0.5, 1.0, 1.5). As shown in Table~\ref{table:Robustness_GenImage} and Table~\ref{table:Robustness_AIGIBench}, experiments on two different datasets reveal that all methods experience performance degradation under perturbations, weakening their ability to reliably distinguish real from fake images. Despite these challenging conditions, our method consistently outperforms competing approaches, maintaining relatively high accuracy. Notably, its performance remains almost unaffected under varying levels of Gaussian blur, further demonstrating the strong robustness of our approach.

\subsection{Comparative Studies of Module Design} 
To further verify that our proposed method helps the model extract more fine-grained multi-angle comprehensive forgery traces while modeling the importance and implicit dependencies between local features, we conducted further comparative studies on the modules, as follows.

1) \textbf{The reconstructed sliding window attention mechanism greatly improves the ability of attention to capture forgery features.} As shown in Table \ref{tab_RSWAttention}, compared with WAttention, our RSWAttention improves the mean Acc. by 24.2\%, which suggests that the RSWAttention greatly enhances the attention mechanism to capture fine-grained forgery traces within the local window while modeling the importance and interdependence between local features. 

2) \textbf{Utilizing only the phase component of the FFT can reduce features not related to forgery compared to using the full FFT.} As shown in Table \ref{tab_fft}, if we use not only the phase part of the FFT but also the amplitude part as input to the network, the average accuracy of the model drops by 5.2\%, confirming that amplitude introduces noise or irrelevant intensity information that confounds the detector. In contrast, the phase, encoding structural details like edges and shapes, aligns with the artifact-relevant features needed for detection. This finding corroborates prior image processing research, where phase information is prioritized over amplitude for structural analysis, reinforcing our design choice to leverage FFT’s phase as a complementary input branch. In addition, replacing FFT with DCT in the complement branch leads to about a 2\% decrease in mAcc., as shown in Table~\ref{tab_dct}, demonstrating the effectiveness of relying on FFT-based features.

3) \textbf{All subbands of DWT are indispensable}. Each frequency domain subband of DWT has its own unique meaning, the low-frequency part retains the overall structure and semantic background of the image, and the high-frequency part captures subtle texture and edge information. Deleting any subband will reduce the comprehensiveness of the information in the window. As shown in Table \ref{tab_dwt}, whether using only one frequency domain subband or omitting some frequency domain subbands, the performance of the model will degrade. 

4) \textbf{The window size for each module is optimal.} To further validate the effectiveness of our window design, we conduct additional experiments to examine the impact of different window sizes. As shown in Table \ref{tab_window}, any deviation from our chosen configuration results in a noticeable performance drop, which further substantiates the suitability of our design. Specifically, in the FFT-based phase complement branch, each element is derived from the global phase spectrum and thus encodes long-range dependencies and structural relationships among the overall elements. As shown in the first row of table \ref{tab_window}, reducing the attention window restricts interactions to only very local neighborhoods of phase features, weakening the model’s ability to capture global phase relationships and resulting in degraded performance. In contrast, the DWT-based window tiling branch is designed to capture comprehensive and fine-grained spatial-frequency domain information, enabling more detailed feature extraction. After the $2\times 2$ window tiling, each window displays a highly organized structure: the representation encodes both the four spatially adjacent feature values from the original image and the corresponding feature information from four distinct frequency subbands. This alignment encourages attention to capture coherent dependencies across both neighboring spatial locations and frequency components. Enlarging the tiling window or changing the attention window disrupts this carefully constructed local structure, mixing unrelated spatial regions and subbands within the same window. Consequently, the representation within each window becomes less coherent, leading to a marked decrease in detection performance, as observed in Table \ref{tab_window}.

\subsection{Hyperparameter $\lambda$ Ablation Study} 
In order to more comprehensively evaluate the effectiveness of our method, we also conduct ablation experiments on the hyperparameter $\lambda$, which can be shown in Table \ref{tab_lambda}. When our parameter $\lambda$ is set to 0.4 as set in the paper, the best results are achieved on all datasets.

\section{Conclusion}

In this paper, we propose a novel and effective method for AI-generated image detection. Specifically, we designed a reconstructed sliding window attention mechanism, which reconstructs the features within the window and limits the attention to the local window, forcing the attention to extract fine-grained features while modeling the importance and dependencies between internal elements in the local area. In addition, we design a dual frequency branch framework consisting of four frequency bands in the DWT domain and the phase part in the FFT domain to extract richer local artifact features from multiple perspectives. Extensive experiments on 65 diverse generative models strongly demonstrate the generalization capability of our method.

\bibliographystyle{unsrt}
\bibliography{ieee}

\end{document}